\documentclass[lettersize,journal]{IEEEtran}
\usepackage{amsmath,amsfonts}
\usepackage{algorithmic}
\usepackage{algorithm}
\usepackage{array}
\usepackage[caption=false,font=normalsize,labelfont=sf,textfont=sf]{subfig}
\usepackage{textcomp}
\usepackage[table]{xcolor}
\usepackage{multirow}
\usepackage{booktabs}
\usepackage{tabularx}
\usepackage{makecell}
\definecolor{tong}{RGB}{230,240,225}
\definecolor{brown}{RGB}{150,75,0}
\usepackage{stfloats}
\usepackage{url}
\usepackage{verbatim}
\usepackage{graphicx}
\usepackage{cite}
\usepackage[colorlinks,
            linkcolor=blue,
            anchorcolor=blue,
            citecolor=blue]{hyperref}

\usepackage{doi}

\usepackage{orcidlink}

\begin{document}


\title{SemDINO: Foundation Prior-Guided Cross-Temporal Semantic Alignment Network for Remote Sensing Change Detection}

\author{Xinyu Tong~\orcidlink{0009-0003-3476-4850},~\IEEEmembership{Student Member,~IEEE}, Meihua Zhou~\orcidlink{0009-0008-4309-7208}, Jinxiao Sun~\orcidlink{0009-0008-3342-1086}, Zaiyan Zhang~\orcidlink{0009-0001-7376-7101},~\IEEEmembership{Student Member,~IEEE}, Hongruixuan Chen~\orcidlink{0000-0003-0100-4786},~\IEEEmembership{Member,~IEEE}, Lei Wang~\orcidlink{0000-0002-8253-7295}

\thanks{Manuscript submitted Aug xx, 2026. \textit{(Xinyu Tong and Meihua Zhou contribute equally.) (Corresponding author: Meihua Zhou and Lei Wang)}}
\thanks{Xinyu Tong and Lei Wang are with the Xinjiang Institute of Ecology and Geography, Chinese Academy of Sciences, Urumqi 830011, China, and also with the University of Chinese Academy of Sciences, Beijing 100049, China. (e-mail:tongxinyu25@mails.ucas.ac.cn,  egiwang@ms.xjb.ac.cn)}
\thanks{Meihua Zhou is with the University of Chinese Academy of Sciences, Beijing 100049, China. (e-mail: zhoumeihua25@mails.ucas.ac.cn)}
\thanks{Jinxiao Sun is with the School of Computer
Science, Xiangtan University, Xiangtan 411105, China.}
\thanks{Zaiyan Zhang is with the School of Geodesy and Geomatics, Wuhan University, Wuhan 430079, China.}
\thanks{Hongruixuan Chen is with the Graduate School of Frontier Sciences, The University of Tokyo, Kashiwa 277-8561, Japan.}}


\maketitle

\begin{abstract}
Semantic change detection (SCD) in remote sensing aims to identify land-cover transitions between bi-temporal observations while suppressing pseudo-changes caused by illumination variations, seasonal differences, and registration errors. Although Vision Foundation Models (VFMs) provide transferable semantic priors, their application to SCD remains challenging due to the mismatch between foundation-model representations and task-specific spatial features, as well as temporal-order sensitivity. To address these issues, this paper proposes SemDINO, a foundation prior-guided framework that integrates transferable vision foundation model priors with hierarchical convolutional representations for cross-temporal semantic reasoning. Specifically, a Gated Pyramid Fusion (PyFu) module is developed to adaptively combine foundation-model semantics with CNN spatial details while reducing domain noise. A Multi-scale Temporal Bi-directional Transformer (M-TBTT) is introduced to achieve symmetric cross-temporal feature interaction and alleviate temporal-order bias. Furthermore, a Feature Change Enhancement (FeaCE) flow is designed to refine aligned representations and distinguish genuine semantic transitions from pseudo variations. Finally, a multi-branch decoupled prediction head jointly generates change masks, bi-temporal semantic maps, and edge constraints. Extensive experiments across five benchmark datasets demonstrate that SemDINO consistently outperforms state-of-the-art methods on both semantic and binary change detection tasks. The results validate the effectiveness of alignment-oriented representation learning for robust remote sensing change analysis. The code will be available in \textcolor{brown}{\textbf{https://github.com/tonxycs/SemDINO}}
\end{abstract}


\section{Introduction}

\IEEEPARstart{S}{emantic} change detection (SCD) is a core research direction of remote sensing image interpretation, which aims to identify fine-grained land-cover conversion types based on bi-temporal image pairs, typically denoted as $T_1$ and $T_2$. Compared with conventional binary change detection (BCD) that only distinguishes changed and unchanged pixels, SCD further provides explicit semantic categories of land cover before and after changes, which supports various practical applications including ecological monitoring, natural disaster response, and urban land management. To facilitate the advancement of SCD research, multiple authoritative public benchmarks have been constructed by the community. Landsat-SCD \cite{Yuan_et_al._2022} provides abundant semantic transition annotations to standardize large-scale, long-term SCD assessment, while SECOND \cite{Yang_et_al._2021} pushes forward high-resolution aerial SCD research with fine-grained category labels for complex surface landscapes. Meanwhile, HRSCD \cite{Daudt_et_al._2019} delivers compact, high-quality annotated samples suitable for model testing and ablation analysis. Collectively, these benchmarks demonstrate that modern SCD is essentially a cross-temporal semantic reasoning task, where reliable transition understanding depends on maintaining semantic consistency between observations acquired at different times under diverse external disturbances.

Despite the progress achieved by existing SCD frameworks, robust semantic transition reasoning remains challenging because bi-temporal observations contain both invariant semantic structures and meaningful temporal evolutions. Unlike binary change detection that only requires locating discrepancy regions, SCD must determine whether the observed difference originates from genuine land-cover transition or from temporal observation noise. Therefore, an ideal SCD representation should simultaneously preserve semantic consistency in unchanged regions and maintain discriminative transition cues in changed regions. Formally, unchanged regions should satisfy reduced cross-temporal representation discrepancy, whereas changed regions should retain category-dependent semantic differences.

Existing SCD frameworks have attempted to address this challenge through different representation learning paradigms. Early post-classification comparison strategies generate semantic change maps by comparing two independent semantic prediction results, but they suffer from severe error propagation because semantic errors from either temporal prediction can directly affect the final transition results. More recent end-to-end multi-task learning frameworks jointly optimize semantic segmentation and change detection objectives, enabling direct cross-temporal semantic reasoning. Representative works such as HRSCD \cite{Daudt_et_al._2019}, BiSRNet \cite{Ding_et_al._2022}, ChangeMask \cite{Zheng_et_al._2022}, SSCD \cite{Ding_et_al._2022}, SCanNet \cite{Ding_et_al._2024}, ChangeMamba \cite{Chen_et_al._2024}, BT-SCD \cite{Tang_et_al._2025}, TaCo \cite{Guo_et_al._2025} and CdSC \cite{Wang_et_al._2024a} improve SCD performance through multi-task semantic reasoning, bi-temporal feature interaction, boundary-aware optimization, and high-level semantic relationship modeling. However, these methods mainly learn semantic representations from limited in-domain SCD annotations. Without generalized semantic priors, the learned representations remain vulnerable to cross-scene variations, temporal disturbances, and complex land-cover transitions.

Visual foundation models (VFMs) provide a promising opportunity to enhance SCD by introducing transferable dense semantic priors learned from large-scale pre-training. Self-supervised visual backbones (e.g., DINO series) learn robust, high-level semantic representations from massive open-world data, offering strong semantic abstraction capabilities for complex remote sensing scenarios. Several recent studies have explored the potential of foundation models for change analysis: ChangeCLIP \cite{Dong_et_al._2024} utilizes vision-language alignment priors, Semantic-CD \cite{Zhu_et_al._2025} realizes open-vocabulary SCD, and VFM-ReSCD \cite{Zhang_et_al._2025a} adapts pre-trained models for high-resolution recurrent change detection. Nevertheless, directly transferring foundation-model representations cannot fully satisfy SCD requirements. Most existing VFM-based approaches primarily focus on binary change localization or category-agnostic mask generation, lacking comprehensive bi-temporal semantic state modeling and fine-grained category transition reasoning. Moreover, foundation representations are acquired from general visual distributions rather than cross-temporal semantic alignment objectives, leading to domain mismatches with task-specific hierarchical CNN representations in spatial resolution and semantic granularity.

\begin{figure}[]
    \centering
    \includegraphics[width=0.45\textwidth]{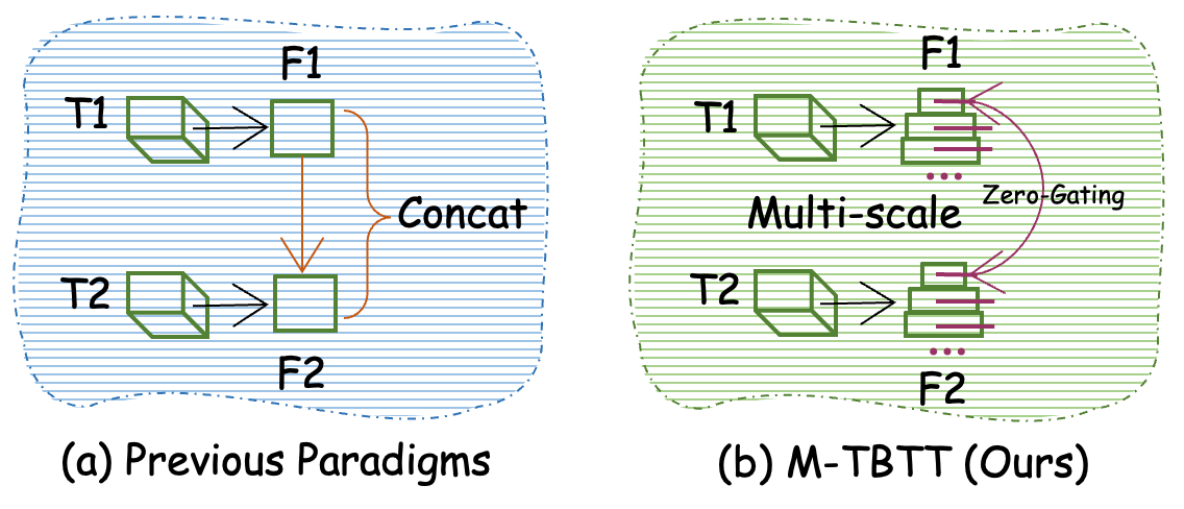}
    \caption{Comparison of cross-temporal feature interaction mechanisms in semantic change detection. (a) Existing paradigms rely on asymmetric feature interaction, including late fusion and directional cross-attention, which may introduce temporal-order sensitivity and insufficient hierarchical alignment. (b) SemDINO utilizes the Multi-scale Bidirectional Temporal Transformer (M-TBTT) to calibrate DINOv3-enhanced bi-temporal representations through symmetric cross-temporal interaction and identity-preserved zero-gating.}
    \label{fig:mtbtt}
\end{figure}

To effectively harness visual foundation priors for SCD, a dedicated cross-temporal calibration mechanism is required to adapt dense foundation representations according to temporal semantic relationships. As illustrated in Fig.~\ref{fig:mtbtt}(a), early feature interaction strategies often rely on asymmetric feature aggregation or direction-dependent interaction schemes, such as feature concatenation and temporal differencing, which fail to explicitly model mutual temporal dependencies. Although pioneering works such as ChangeMask \cite{Zheng_et_al._2022} and Changer \cite{Fang_et_al._2023} have highlighted the importance of temporal exchangeability, their interactions are primarily constrained to standard CNN features via basic feature differencing or shallow cross-attention mechanisms. Consequently, conventional asymmetric or shallow temporal designs may compromise the inherent temporal exchangeability of observations, making predictions sensitive to the input order of $T_1$ and $T_2$. Furthermore, many existing methods lack joint calibration across multi-scale feature pyramids, where high-level semantic abstractions and low-level spatial details are inadequately synchronized. As a result, unchanged regions remain vulnerable to pseudo changes caused by insufficient semantic alignment, whereas subtle real transitions are easily suppressed by excessive feature smoothing.

To overcome these limitations and enable effective utilization of DINOv3-enhanced representations, the Multi-scale Bidirectional Temporal Transformer (M-TBTT) is developed as the core cross-temporal calibration mechanism. M-TBTT reformulates cross-temporal interaction as a symmetric semantic calibration process rather than one-way feature matching. As shown in Fig.~\ref{fig:mtbtt}(b), M-TBTT constructs two reciprocal temporal interaction streams that update both observations through mutual conditioning. Specifically, the $T_1\rightarrow T_2$ and $T_2\rightarrow T_1$ branches simultaneously calibrate temporal representations, enabling observation-induced discrepancies to be reduced while transition-specific semantic differences are preserved. Furthermore, M-TBTT performs alignment across hierarchical feature pyramids, allowing shallow representations to retain boundary details and deep representations to maintain semantic consistency. The identity-preserved zero-gating mechanism initializes the module as an unchanged mapping and gradually introduces temporal correction during optimization, thereby preventing destructive disturbance to pretrained semantic representations. Through symmetric and progressive cross-temporal calibration, M-TBTT establishes an effective bridge between DINOv3 semantic priors and SCD-oriented transition reasoning.

Based on this principle, SemDINO is proposed as a foundation prior-guided cross-temporal semantic alignment framework for remote sensing change detection. First, a Pyramid Fusion (PyFu) module is designed to adapt frozen DINOv3 representations and integrate foundation-model semantic priors with hierarchical CNN features. Subsequently, M-TBTT performs symmetric cross-temporal calibration on the hybrid representations, enabling DINOv3 semantic knowledge to be effectively transferred into temporal transition reasoning. After semantic alignment, a Feature Change Enhancement (FeaCE) pipeline composed of bidirectional change enhancement (BCE), semantic clean purification (SCP), and multi-scale change enhancement (MCE) is introduced to refine transition-specific representations and suppress pseudo variations. Finally, a decoupled multi-task Change Detection Head (CD-Head) is constructed to simultaneously generate binary change maps, dual-temporal semantic maps, and edge guidance maps. Through the joint optimization of semantic prior adaptation and cross-temporal calibration, SemDINO provides a unified solution for exploiting foundation-model knowledge in SCD. Extensive experiments conducted on Landsat-SCD, SECOND, and HRSCD datasets demonstrate the effectiveness and generalization capability of the proposed framework.

The main contributions of this work are summarized as follows:

1) We propose SemDINO, a novel DINOv3-guided cross-temporal semantic alignment framework for remote sensing change detection. By leveraging open-world semantic priors from a frozen DINOv3 as the core semantic driver, SemDINO establishes a co-evolutionary loop between spatial semantic priors and cross-temporal alignment dynamics.

2) The Multi-scale Bidirectional Temporal Transformer (M-TBTT) is developed as the core cross-temporal calibration mechanism. Symmetric temporal interaction and identity-preserved zero-gating are introduced to adapt DINOv3-enhanced representations while maintaining unchanged-region consistency and changed-region discrimination.

3) A complete refinement and prediction pipeline based on FeaCE and CD-Head is constructed. Multi-scale transition enhancement and decoupled multi-task prediction are incorporated to suppress pseudo changes and support unified SCD and BCD inference.

\section{Related Work}

\subsection{Semantic Change Detection in Remote Sensing}

SCD aims to identify not only where land-cover changes occur but also which semantic categories are involved before and after the transition. Compared with BCD, SCD places a stronger requirement on semantic interpretation, as the model must jointly infer the change mask and the two temporal semantic maps. Early solutions often relied on post-classification comparison, where two independently predicted semantic maps are compared to obtain semantic transitions. This pipeline is intuitive, but classification errors at either temporal image can be directly propagated into the final change map. Recent SCD methods therefore formulate the problem as a joint learning task that couples change localization with bi-temporal semantic parsing. HRSCD introduces large-scale semantic change detection with multitask learning, and BiSRNet further models bi-temporal semantic reasoning for high-resolution SCD \cite{Daudt_et_al._2019}, \cite{Ding_et_al._2022}, \cite{Niu_et_al._2023}. Subsequent methods strengthen the interaction between temporal semantics and change cues through spatio-temporal modeling, state-space representation, or task interaction \cite{Ding_et_al._2024}, \cite{Chen_et_al._2024}, \cite{Tang_et_al._2025}. Other recent SCD studies further investigate cross-difference semantic consistency, semantic-change relationship modeling, decoder-focused prediction, high-resolution feature decoding, and late-stage bi-temporal fusion \cite{Wang_et_al._2024a}, \cite{Tang_et_al._2024}, \cite{Li_et_al._2024}, \cite{Fang_et_al._2024}, \cite{Zhou_et_al._2024}. Despite these advances, SCD still faces a difficult coupling between category-agnostic change localization and category-specific semantic recognition. The change branch tends to respond to temporal discrepancy, while the semantic branches require stable land-cover representations across time. Under illumination imbalance, seasonal vegetation phenological variations, and minor registration errors, this mismatch may suppress true semantic transitions or amplify pseudo changes. Therefore, robust SCD relies heavily on the ability to maintain semantic consistency in unchanged regions and accurately distinguish category-level transitions in changed regions.

\subsection{Multi-scale Semantic Representation Learning for SCD}

High-resolution SCD requires multi-scale semantic representations because land-cover transitions range from small object-level changes to large region-level conversions. Shared bi-temporal encoders and feature pyramids have been widely adopted to extract comparable representations from the two temporal images, while multi-level decoders are used to recover dense semantic and change predictions. In SCD, however, multi-scale learning is more challenging than in binary change detection. Shallow features preserve boundaries and small structures, but they are sensitive to radiometric disturbance and minor registration deviations. Deep features encode stronger semantics, but they may lose fine spatial details that are critical for precise semantic transitions. Existing SCD methods alleviate this tradeoff through bi-temporal semantic reasoning, joint spatio-temporal modeling, boundary-aware learning, task interaction, dual-dimension feature interaction, and semantic enhancement with change consistency constraints \cite{Ding_et_al._2022}, \cite{Ding_et_al._2024}, \cite{Tang_et_al._2025}, \cite{Wang_et_al._2024b}, \cite{Jiang_et_al._2025}. Broader remote sensing representation learning has also explored self-supervised pretraining, multi-spectral or multi-scale reconstruction, and spectral curriculum strategies to improve transferable feature quality \cite{Manas_et_al._2021}, \cite{Cong_et_al._2022}, \cite{Noman_et_al._2024}, \cite{Reed_et_al._2022}, \cite{Zhou_et_al._2025}, \cite{Zhang_et_al._2025c}. These designs confirm that SCD performance relies on whether multi-scale features can effectively support temporal semantic consistency and change discrimination. Nevertheless, most existing SCD frameworks learn semantic representations only from task-specific remote sensing annotations. Due to the limited scene diversity, semantic coverage, and annotation density of public datasets, the learned pyramidal features remain vulnerable to cross-scene domain shifts and pseudo temporal variations.

\subsection{Foundation-model Priors for Semantic Change Detection}

Visual foundation models (VFMs) offer highly transferable dense representations for semantic change detection (SCD) through large-scale pretraining. Self-supervised models such as DINO, DINOv2, and DINOv3 provide strong semantic abstraction without manual labels \cite{Caron_et_al._2021}, \cite{Oquab_et_al._2024}, \cite{Simeoni_et_al._2025}, while multimodal and segmentation foundation models like CLIP and SAM demonstrate the value of large-scale pretraining for open-set semantics and transferable dense prediction \cite{Radford_et_al._2021}, \cite{Kirillov_et_al._2023}. In remote sensing, domain-adapted models such as RemoteCLIP, RingMo-Aerial, SkySense, and Prithvi extend visual and vision-language pretraining to Earth observation data \cite{Liu_et_al._2024}, \cite{Diao_et_al._2024}, \cite{Guo_et_al._2023}, \cite{Jakubik_et_al._2023}.

For change analysis, ChangeCLIP exploits vision-language priors \cite{Dong_et_al._2024}, and ChangeDINO introduces frozen DINOv3 features into a multi-scale Siamese framework for building change detection \cite{Cheng_et_al._2026}. Recent works have further integrated visual and segmentation foundation models into SCD tasks \cite{Mei_et_al._2024}, \cite{Shen_et_al._2026}, \cite{Huang_et_al._2026}.

Although these studies confirm that foundation priors improve model robustness under limited annotations and imaging variations, key challenges remain. Most existing VFM-adapted methods focus on binary or object-specific change localization, where pretrained features are primarily used to refine change boundaries. In contrast, SCD requires precise bi-temporal semantic state estimation and reasoning over complex "from-to" transitions. Furthermore, pretrained dense representations differ substantially from standard task-specific CNN feature pyramids in scale, channel dimensionality, and feature distributions. How to effectively adapt and align foundation priors into multi-scale, temporally consistent semantic representations for SCD remains insufficiently explored.

\begin{figure*}[htbp]
    \centering
    \includegraphics[width=\textwidth]{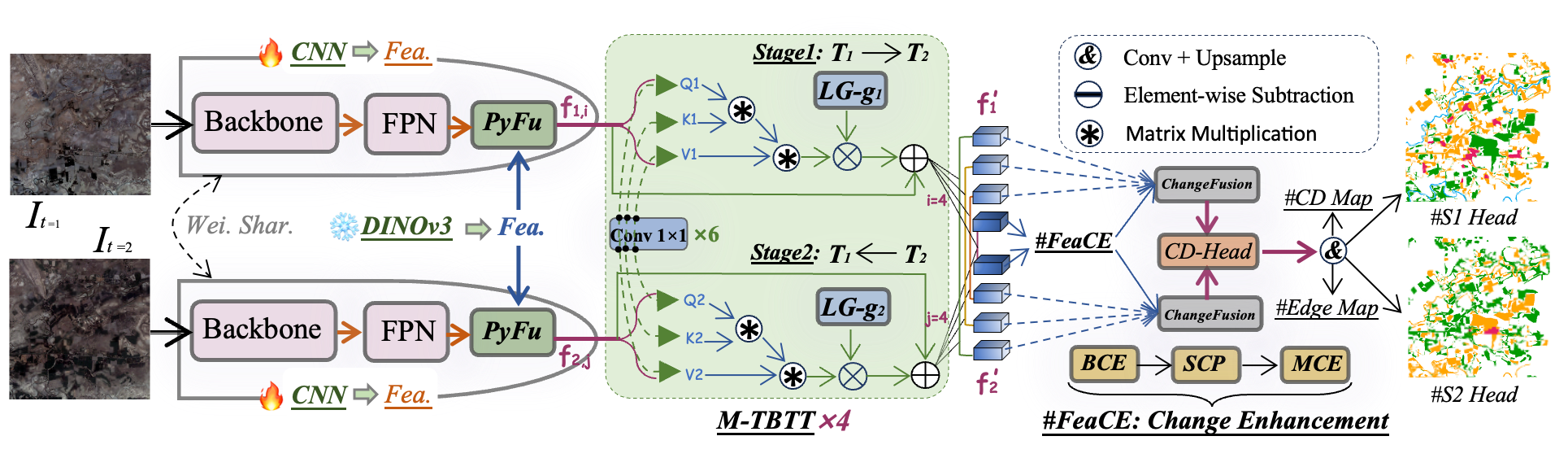}
    \caption{\small Overview of the proposed SemDINO framework.
Given bi-temporal remote sensing images $I_{t=1}$ and $I_{t=2}$, the network first extracts multi-scale features using a CNN backbone with FPN, and enhances them with complementary features from the frozen \textbf{\textit{DINOv3}} encoder. The Pyramid Fusion (\textbf{\textit{PyFu}}) module then fuses the CNN and DINO features at each scale. Next, the Multi-scale Bidirectional Temporal Transformer (\textbf{\textit{M-TBTT}}) aligns the bi-temporal features in both directions ($T_1 \to T_2$ and $T_1 \gets T_2$), with learnable gating (\textit{\textbf{LG-g}}) to adaptively control the alignment strength. After alignment, the change enhancement pipeline \#\textit{\textbf{FeaCE}} (BCE, SCP, MCE) refines the change features, which are then fused by two parallel \textbf{\textit{ChangeFusion}} modules. Finally, the \textbf{multi-task CD-Head} simultaneously outputs the change detection (CD) map, the semantic segmentation maps for both $T_1$ and $T_2$, and the auxiliary edge map for supervision. }
    \label{fig:main}
\end{figure*}

\section{Methodology}

\subsection{Overview of the SemDINO Framework}

Given two co-registered remote sensing images $I_{t=1}$ and $I_{t=2}$, SCD aims to predict a semantic change map that describes both change localization and semantic transition. In SemDINO, this target is constructed through a decomposed output scheme, including CD Map, S1 Head, S2 Head, and Edge Map. The CD Map localizes changed pixels. S1 Head and S2 Head predict the semantic states at $T_1$ and $T_2$. Edge Map provides auxiliary boundary supervision.

For a pixel $p$, the semantic transition relation is written as
\begin{equation}
Y_{\mathrm{scd}}(p)
=
\mathcal{H}
\left(
Y_{\mathrm{cd}}(p),
Y_{\mathrm{s1}}(p),
Y_{\mathrm{s2}}(p)
\right),
\end{equation}
where $\mathcal{H}(\cdot)$ denotes the SCD composition rule. When $Y_{\mathrm{cd}}(p)=0$, the pixel is assigned to the unchanged class. When $Y_{\mathrm{cd}}(p)=1$, the semantic transition is determined by the semantic states at $T_1$ and $T_2$. This relation gives two requirements. Unchanged regions require cross-temporal semantic consistency. Changed regions require semantic discrepancy to be preserved and converted into from-to transition categories.

As shown in Fig.~\ref{fig:main}, \textbf{the technical core of SemDINO revolves around a co-evolutionary loop of cross-temporal semantic alignment, where spatial semantic priors and temporal alignment dynamics mutually reinforce each other}. Specifically, PyFu establishes the foundation by injecting open-world semantic embeddings from a frozen DINOv3 into CNN features. These robust spatial priors drive the subsequent alignment process, preventing the model from drifting into low-level pixel noise. Upon this semantic manifold, M-TBTT executes bidirectional temporal alignment through Stage 1 ($T_1\rightarrow T_2$) and Stage 2 ($T_1\leftarrow T_2$), which in turn guides and calibrates the static DINO features to capture cross-temporal evolutionary dynamics. This reciprocal interaction is central to SCD, ensuring that the bi-temporal semantic states are fully aligned before their discrepancy is decoded as a meaningful semantic transition. Following this core alignment loop, downstream auxiliary components are deployed for feature refinement and final prediction: FeaCE serves as a purification filter to suppress pseudo variations, while ChangeFusion integrates the aligned features to feed the decoupled CD-Head, which outputs the CD Map, S1/S2 Heads, and Edge Map for comprehensive SCD composition.

For compact notation, $F_{\mathrm{pyramid},t}^{i}$ and $F_{\mathrm{dino},t}^{i}$ are denoted by $F_{t,i}^{p}$ and $F_{t,i}^{d}$, respectively. For each temporal image $I_t$, feature extraction and fusion are written as
\begin{equation}
\begin{aligned}
F_{t,i}^{p}
&=
\mathrm{FPN}_{i}(\mathrm{Backbone}(I_t)),\\
F_{t,i}^{d}
&=
\mathrm{SepAB}_{i}(\mathrm{DINOv3}(I_t)),\\
f_{t,i}
&=
\mathrm{PyFu}_{i}(F_{t,i}^{p},F_{t,i}^{d}),
\end{aligned}
\quad
t\in\{1,2\},\ i=1,\ldots,4.
\end{equation}

M-TBTT performs bidirectional temporal interaction at each pyramid level. To keep the formulation compact, the cross-temporal attention operator is denoted by
\begin{equation}
\mathcal{A}(Q,K,V)
=
\mathrm{Softmax}
\left(
\frac{QK^{\top}}{\sqrt{d}}
\right)V .
\end{equation}
For the $i$-th level, Stage1: $T_1\rightarrow T_2$ is formulated as
\begin{equation}
\begin{aligned}
O_{1\rightarrow2,i}
&=
\mathcal{A}
\left(
W_{q1}f_{1,i},
W_{k1}f_{2,i},
W_{v1}f_{2,i}
\right),\\
f'_{1,i}
&=
f_{1,i}
+
g_{1,i} \odot O_{1\rightarrow2,i}.
\end{aligned}
\end{equation}
Stage2: $T_1\leftarrow T_2$ is symmetric:
\begin{equation}
\begin{aligned}
O_{2\rightarrow1,i}
&=
\mathcal{A}
\left(
W_{q2}f_{2,i},
W_{k2}f_{1,i},
W_{v2}f_{1,i}
\right),\\
f'_{2,i}
&=
f_{2,i}
+
g_{2,i} \odot O_{2\rightarrow1,i}.
\end{aligned}
\end{equation}

By strictly sharing the projection parameters $(W_q, W_k, W_v)$ and the adaptive gating factor $g_i$ (implemented via LG-g) across both temporal branches, M-TBTT mathematically guarantees permutation equivariance, ensuring that swapping the temporal inputs $T_1$ and $T_2$ yields strictly symmetric feature updates. Furthermore, by initializing $g_i$ to zero, M-TBTT defaults to an identity mapping at initialization, allowing cross-temporal corrections to be learned gradually while maintaining optimization stability during early training stages.

More importantly, M-TBTT differs fundamentally from naive feature differencing, concatenation, and one-way temporal attention mechanisms, as it avoids treating either timestamp as an absolute static reference. Instead, it dynamically updates both temporal representations under mutual cross-conditioning: the $T_1\rightarrow T_2$ path calibrates $f_{1,i}$ with respect to $f_{2,i}$, whereas the $T_2\rightarrow T_1$ path calibrates $f_{2,i}$ with respect to $f_{1,i}$. Owing to this strict parameter sharing and structural symmetry, M-TBTT effectively eliminates temporal-order bias and establishes a robust semantic foundation for distinguishing invariant background regions from valid land-cover transitions. \textbf{The comprehensive mathematical formalization, proof of bounded temporal-order bias, and gradient stability upper-bound regarding this bidirectional alignment structure are thoroughly documented in Sections I-1.2 and I-1.3 of the Supplementary Material.}

The aligned deep features are sent to FeaCE:
\begin{equation}
d
=
F_{\mathrm{FeaCE}}
=
\mathrm{FeaCE}(f'_{1,4},f'_{2,4}).
\end{equation}
Following Fig.~\ref{fig:feace}, $d$ denotes the enhanced change feature generated by BCE, SCP, and MCE. Let
\begin{equation}
\mathcal{S}
=
\{
f'_{1,1},f'_{1,2},f'_{1,3},
f'_{2,1},f'_{2,2},f'_{2,3},d
\}
\end{equation}
denote the feature set used by ChangeFusion. It contains three aligned features from $T_1$, three aligned features from $T_2$, and one enhanced change feature. All features are resized to the same spatial resolution and concatenated along the channel dimension:
\begin{equation}
X
=
\mathrm{Concat}
\left(
\left\{
\mathcal{U}(f)\mid f\in\mathcal{S}
\right\}
\right),
\end{equation}
where $\mathcal{U}(\cdot)$ denotes bilinear upsampling. Since $\mathcal{S}$ contains $3+3+1$ groups of features, $X$ aggregates seven feature groups.

\textbf{ChangeFusion is formulated as}
\begin{equation}
F_{\mathrm{scd}}
=
\mathrm{ChangeFusion}(f'_{1,1:3},f'_{2,1:3},d)
=
\mathcal{F}(X)
\odot
\mathrm{CA}
\left(
\mathcal{F}(X)
\right),
\end{equation}
where $\mathcal{F}(\cdot)$ denotes the $1\times1$ convolutional fusion layer, $\mathrm{CA}(\cdot)$ denotes the channel attention recalibration in ChangeFusion, and $\odot$ denotes element-wise multiplication. The multi-task CD-Head predicts the decomposed outputs:
\begin{equation}
(\hat{Y}_{\mathrm{cd}},
\hat{Y}_{\mathrm{s1}},
\hat{Y}_{\mathrm{s2}},
\hat{Y}_{\mathrm{edge}})
=
\mathrm{CD\mbox{-}Head}(F_{\mathrm{scd}}).
\end{equation}

The final semantic change map is composed as
\begin{equation}
\hat{Y}_{\mathrm{scd}}(p)=
\begin{cases}
0, & \sigma(\hat{Y}_{\mathrm{cd}}(p))<\tau,\\
\Psi\left(
\arg\max \hat{Y}_{\mathrm{s1}}(p),
\arg\max \hat{Y}_{\mathrm{s2}}(p)
\right), & \sigma(\hat{Y}_{\mathrm{cd}}(p))\geq\tau,
\end{cases}
\end{equation}
where $\tau$ is the change threshold and $\Psi(\cdot)$ denotes the from-to semantic transition encoding.

The training objective supervises the decomposed outputs:
\begin{equation}
\begin{aligned}
\mathcal{L}
&=
\lambda_{\mathrm{cd}}\mathcal{L}_{\mathrm{cd}}
+
\lambda_{\mathrm{s1}}\mathcal{L}_{\mathrm{s1}}
+
\lambda_{\mathrm{s2}}\mathcal{L}_{\mathrm{s2}}
+
\lambda_{\mathrm{edge}}\mathcal{L}_{\mathrm{edge}}
+
\lambda_{\mathrm{sc}}\mathcal{L}_{\mathrm{sc}},\\
\mathcal{L}_{\mathrm{cd}}
&=
\mathrm{WeightedBCE}(\hat{Y}_{\mathrm{cd}},Y_{\mathrm{cd}}),\\
\mathcal{L}_{\mathrm{s1}}
&=
\mathrm{CE}(\hat{Y}_{\mathrm{s1}},Y_{\mathrm{s1}}),
\quad
\mathcal{L}_{\mathrm{s2}}
=
\mathrm{CE}(\hat{Y}_{\mathrm{s2}},Y_{\mathrm{s2}}),\\
\mathcal{L}_{\mathrm{edge}}
&=
\mathrm{BCE}(\hat{Y}_{\mathrm{edge}},Y_{\mathrm{cd}}),\\
\mathcal{L}_{\mathrm{sc}}
&=
\mathrm{ChangeSimilarity}(\hat{Y}_{\mathrm{s1},1:},\hat{Y}_{\mathrm{s2},1:},Y_{\mathrm{cd}}).
\end{aligned}
\end{equation}
We set fixed weighting coefficients following the training pipeline: $\lambda_{\mathrm{cd}}=1$, $\lambda_{\mathrm{s1}}=0.5$, $\lambda_{\mathrm{s2}}=0.5$, $\lambda_{\mathrm{edge}}=0.1$, $\lambda_{\mathrm{sc}}=1$.
The five loss terms collaboratively optimize the final SCD objective. The CD branch judges whether a pixel undergoes a semantic transition. The S1 Head and S2 Head predict per-pixel land-cover semantic states for bi-temporal images respectively. The edge loss regularizes ambiguous boundary regions where semantic transitions are prone to misjudgment. Additionally, $\mathcal{L}_{\mathrm{sc}}$ acts as a similarity regularizer, constraining semantic feature consistency within unchanged pixels to suppress false positive change noise. 

\textit{(Note: $\operatorname{CE}$/$\operatorname{BCE}$/$\operatorname{WeightedBCE}$ are loss functions, different from the Bi-Change Enhancement (BCE) module in FeaCE. $\operatorname{WeightedBCE}$ adopts 0.25/0.75 pos/neg weights for class imbalance. $\operatorname{ChangeSimilarity}$ uses cosine embedding to reduce false changes.)}

\begin{algorithm}
\caption{SemDINO Optimization}
\label{alg:semdino}
\begin{algorithmic}[1]
\REQUIRE Images $I_{t=1},I_{t=2}$; labels $Y_{\mathrm{cd}},Y_{\mathrm{s1}},Y_{\mathrm{s2}},Y_{\mathrm{edge}}$; parameters $\Theta$
\ENSURE $\hat{Y}_{\mathrm{cd}}$, $\hat{Y}_{\mathrm{s1}}$, $\hat{Y}_{\mathrm{s2}}$, $\hat{Y}_{\mathrm{edge}}$, and $\hat{Y}_{\mathrm{scd}}$
\STATE Compute $F_{1,i}^{p}$ and $F_{2,i}^{p}$ by Backbone+FPN.
\STATE Compute $F_{1,i}^{d}$ and $F_{2,i}^{d}$ by frozen DINOv3 and SepAB.
\FOR{$i=1$ to $4$}
    \STATE Fuse features by PyFu to obtain $f_{1,i}$ and $f_{2,i}$.
    \STATE Compute Stage1 response $O_{1\rightarrow2,i}$ and update $f'_{1,i}=f_{1,i}+g_{1,i}O_{1\rightarrow2,i}$.
    \STATE Compute Stage2 response $O_{2\rightarrow1,i}$ and update $f'_{2,i}=f_{2,i}+g_{2,i}O_{2\rightarrow1,i}$.
\ENDFOR
\STATE Compute $d=F_{\mathrm{FeaCE}}$ by BCE, SCP, and MCE.
\STATE Build $X=\mathrm{Concat}(\{\mathcal{U}(f)\mid f\in\mathcal{S}\})$.
\STATE Compute $F_{\mathrm{scd}}=\mathcal{F}(X)\odot\mathrm{CA}(\mathcal{F}(X))$.
\STATE Predict CD Map, S1 Head, S2 Head, and Edge Map by the multi-task CD-Head.
\STATE Compose $\hat{Y}_{\mathrm{scd}}$ from CD Map, S1 Head, and S2 Head.
\STATE Minimize $\mathcal{L}$ and update $\Theta$.
\end{algorithmic}
\end{algorithm}

\subsection{Pyramid Fusion (PyFu)}

\begin{figure}[]
\centering
\includegraphics[width=0.48\textwidth]{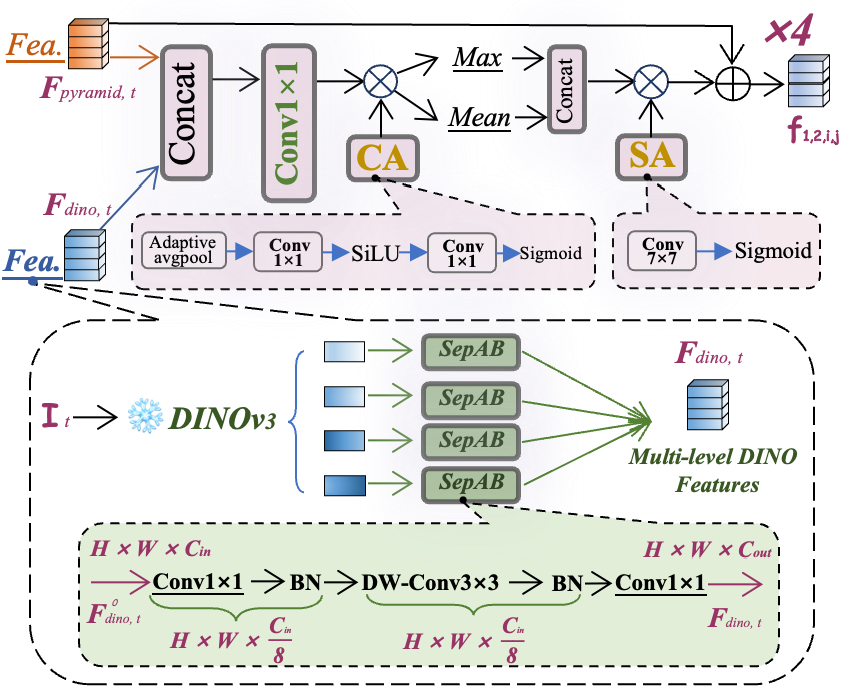}
\caption{{\small Overview of the Pyramid Fusion (\textbf{PyFu}) module and multi-level feature extraction from DINOv3.
Given the input image $I_t$, the frozen DINOv3 encoder extracts multi-level semantic features, which are then processed by \textbf{Separate Adaptation Blocks (SepAB)} to generate aligned multi-level DINO features $F_{dino,t}$. Each SepAB adapts the DINO features via a bottleneck structure: Conv$1\times1$ $\rightarrow$ BN $\rightarrow$ depth-wise Conv$3\times3$ $\rightarrow$ BN $\rightarrow$ Conv$1\times1$. At each pyramid scale, the aligned DINO features $F_{dino,t}$ are concatenated with the CNN pyramid features $F_{pyramid,t}$. The concatenated features are then fed into our \textbf{GatedFusion} module, which first projects them via a $1\times1$ convolution, then refines them through a \textbf{Channel Attention (CA)} branch and a \textbf{Spatial Attention (SA)} branch. The CA branch uses adaptive average pooling and a gated convolution structure to enhance channel-wise dependencies, while the SA branch employs a $7\times7$ convolution to capture spatial context. The outputs of the two attention branches are aggregated and combined with a residual connection to produce the final enhanced features $f_{t,i}$. The entire PyFu module is applied at four pyramid levels ($\times 4$) to obtain multi-scale fused features.}}
\label{fig:Vis2}
\end{figure}

PyFu is designed for semantic-prior injection. Its structure is shown in Fig.~\ref{fig:Vis2}. CNN pyramid features preserve local textures, boundaries, and small structures. However, their semantic abstraction depends heavily on task-specific SCD labels. DINOv3 features provide stronger semantic context, but they are not naturally aligned with CNN pyramid features in scale, channel dimension, and feature distribution. PyFu addresses this mismatch by using SepAB for adaptation and GatedFusion for selective residual fusion.

For the $i$-th pyramid level, SepAB adapts the DINOv3 feature. Then, the adapted DINOv3 feature and the CNN pyramid feature are concatenated and projected:
\begin{equation}
\begin{aligned}
\tilde{F}_{t,i}^{d}
&=
\mathrm{SepAB}_{i}(F_{t,i}^{d}),\\
Z_{t,i}
&=
\phi_{1\times1}
\left(
[
F_{t,i}^{p},
\tilde{F}_{t,i}^{d}
]
\right).
\end{aligned}
\end{equation}
SepAB adopts a bottleneck form:
\begin{equation}
\mathrm{SepAB}:
\mathrm{Conv}_{1\times1}
\rightarrow
\mathrm{BN}
\rightarrow
\mathrm{DW\mbox{-}Conv}_{3\times3}
\rightarrow
\mathrm{BN}
\rightarrow
\mathrm{Conv}_{1\times1}.
\end{equation}

The CA and SA branches produce channel and spatial gates:
\begin{equation}
\begin{aligned}
G_{t,i}^{\mathrm{ca}}
&=
\sigma
\left(
\phi_{c2}
\left(
\delta(
\phi_{c1}(\mathrm{GAP}(Z_{t,i}))
)
\right)
\right),\\
Z_{t,i}^{\mathrm{ca}}
&=
Z_{t,i}\odot G_{t,i}^{\mathrm{ca}},\\
G_{t,i}^{\mathrm{sa}}
&=
\sigma
\left(
\phi_{7\times7}
\left(
[
\mathrm{Avg}_{c}(Z_{t,i}^{\mathrm{ca}}),
\mathrm{Max}_{c}(Z_{t,i}^{\mathrm{ca}})
]
\right)
\right),\\
Z_{t,i}^{\mathrm{sa}}
&=
Z_{t,i}^{\mathrm{ca}}\odot G_{t,i}^{\mathrm{sa}}.
\end{aligned}
\end{equation}
Here, $\mathrm{GAP}(\cdot)$ denotes adaptive average pooling, $\delta(\cdot)$ denotes SiLU, and $\mathrm{Avg}_{c}(\cdot)$ and $\mathrm{Max}_{c}(\cdot)$ denote channel-wise average pooling and max pooling. The final PyFu output is
\begin{equation}
f_{t,i}
=
F_{t,i}^{p}
+
Z_{t,i}^{\mathrm{sa}}.
\end{equation}

The role of PyFu can be stated from semantic error decomposition. Let $S_t^i$ denote an ideal SCD-oriented semantic feature at level $i$. A CNN pyramid feature is written as
\begin{equation}
F_{t,i}^{p}
=
S_t^i
+
\epsilon_{t,i}^{\mathrm{loc}}
+
\epsilon_{t,i}^{\mathrm{sem}},
\end{equation}
where $\epsilon_{t,i}^{\mathrm{loc}}$ denotes local disturbance and $\epsilon_{t,i}^{\mathrm{sem}}$ denotes semantic insufficiency. PyFu injects a DINOv3-based residual compensation:
\begin{equation}
\begin{aligned}
\Delta_{t,i}
&=
Z_{t,i}
\odot
G_{t,i}^{\mathrm{ca}}
\odot
G_{t,i}^{\mathrm{sa}},\\
f_{t,i}
&=
F_{t,i}^{p}
+
\Delta_{t,i}.
\end{aligned}
\end{equation}
Since CA and SA are generated by sigmoid functions, the gates satisfy
\begin{equation}
0\leq G_{t,i}^{\mathrm{ca}}\leq1,
\quad
0\leq G_{t,i}^{\mathrm{sa}}\leq1.
\end{equation}
Thus, the compensation is controlled:
\begin{equation}
\|\Delta_{t,i}\|
\leq
\left\|
\phi_{1\times1}
\left(
[
F_{t,i}^{p},
\tilde{F}_{t,i}^{d}
]
\right)
\right\|.
\end{equation}
If the DINOv3 prior compensates part of $\epsilon_{t,i}^{\mathrm{sem}}$, the semantic representation error satisfies
\begin{equation}
\begin{aligned}
f_{t,i}-S_t^i
&=
\epsilon_{t,i}^{\mathrm{loc}}
+
\epsilon_{t,i}^{\mathrm{sem}}
+
\Delta_{t,i},\\
\|f_{t,i}-S_t^i\|
&\leq
\|\epsilon_{t,i}^{\mathrm{loc}}\|
+
\|\epsilon_{t,i}^{\mathrm{sem}}+\Delta_{t,i}\|.
\end{aligned}
\end{equation}
This expression explains the design objective. PyFu keeps the local-detail representation of CNN-FPN and adds a gated DINOv3 semantic compensation to reduce semantic insufficiency. For SCD, this is necessary because unchanged-region consistency and changed-region transition discrimination both depend on reliable semantic states at $T_1$ and $T_2$. \textbf{Section I-1.4 of the Supplementary Material provides an error decomposition bound to evaluate the semantic compensation capability of this gated prior injection scheme. Crucially, this bound theoretically justifies why incorporating a frozen DINOv3 prior via adaptive gating outperforms a pure task-specific CNN or a fully fine-tuned foundation model, successfully balancing semantic insufficiency and domain discrepancy.}

\subsection{FeaCE: Change Enhancement Structure}

\begin{figure*}[]
    \centering
    \includegraphics[width=\textwidth]{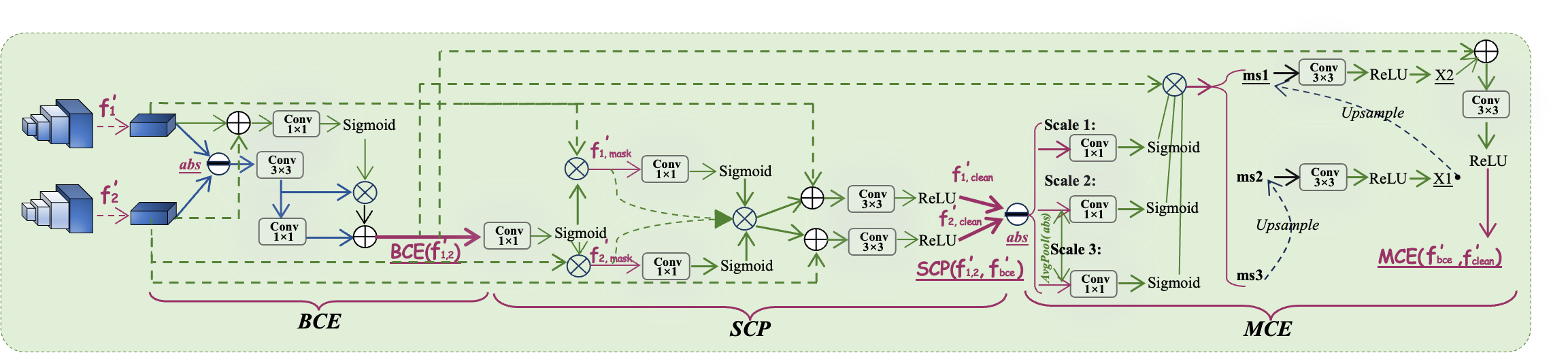}
    \caption{\small Overview of \textbf{FeaCE}: Change Enhancement Structure. Given the aligned bi-temporal features $f_1'$ and $f_2'$, the pipeline consists of three sequential modules: \textbf{a.} \textbf{Bidirectional Change  Enhancement (BCE)} computes the absolute difference of the input features to extract initial change information, which is then enhanced by a learnable gating branch derived from the sum of the two features. A residual convolution branch is added to preserve detailed change cues, producing the initial change feature. \textbf{b.} \textbf{Semantic Clean Purify (SCP)} generates a change mask from the BCE output, and obtains non-change regions by inverting the mask. It then filters the bi-temporal features with the non-change mask, applies learnable gates to the filtered features, and refines them with convolution layers to yield clean bi-temporal features $f_{1,\mathrm{clean}}'$ and $f_{2,\mathrm{clean}}'$. \textbf{c.} \textbf{Multi-scale Change Enhancement (MCE)} first computes the absolute difference of the two clean features, then constructs multi-scale representations via identity mapping, 2× and 4× average pooling. At each scale, attention weights are applied to the pooled change features. These multi-scale features are fused from coarse to fine through bilinear upsampling and convolution layers. Finally, a residual connection with the original BCE output is adopted to generate the final enhanced change feature.}
\label{fig:feace}
\end{figure*}

The FeaCE module is shown in Fig.~\ref{fig:feace}. It receives the aligned deep features $f'_{1,4}$ and $f'_{2,4}$ from M-TBTT. Its objective is to separate unchanged-region semantic consistency from changed-region semantic transition discrimination. It contains BCE, SCP, and MCE.

BCE extracts an initial change feature from the feature difference and shared temporal context:
\begin{equation}
\begin{aligned}
D_{\mathrm{bce}}
&=
\phi_d(|f'_{1,4}-f'_{2,4}|),\\
G_{\mathrm{bce}}
&=
\sigma(\phi_g(f'_{1,4}+f'_{2,4})),\\
F_{\mathrm{BCE}}
&=
D_{\mathrm{bce}}
\odot
G_{\mathrm{bce}}
+
\phi_r(D_{\mathrm{bce}}).
\end{aligned}
\end{equation}
The absolute difference captures temporal discrepancy, the sum-derived gate introduces shared context, and the residual branch preserves detailed change cues.

SCP estimates a change mask and uses its inverse to filter non-change features:
\begin{equation}
\begin{aligned}
M_{\mathrm{chg}}
&=
\sigma(\phi_m(F_{\mathrm{BCE}})),
\quad
M_{\mathrm{nchg}}
=
1-M_{\mathrm{chg}},\\
f'_{1,\mathrm{mask}}
&=
f'_{1,4}\odot M_{\mathrm{nchg}},
\quad
f'_{2,\mathrm{mask}}
=
f'_{2,4}\odot M_{\mathrm{nchg}},\\
G_1
&=
\sigma(\phi_{g1}(f'_{1,\mathrm{mask}})),
\quad
G_2
=
\sigma(\phi_{g2}(f'_{2,\mathrm{mask}})).
\end{aligned}
\end{equation}
The clean temporal features are obtained by gated cross-temporal refinement:
\begin{equation}
\begin{aligned}
f'_{1,\mathrm{clean}}
&=
\phi_1
\left(
f'_{1,4}
+
G_1\odot f'_{2,\mathrm{mask}}
\right),\\
f'_{2,\mathrm{clean}}
&=
\phi_2
\left(
f'_{2,4}
+
G_2\odot f'_{1,\mathrm{mask}}
\right).
\end{aligned}
\end{equation}

Let $\Omega_0=\{p:Y_{\mathrm{cd}}(p)=0\}$ denote unchanged regions and $\Omega_1=\{p:Y_{\mathrm{cd}}(p)=1\}$ denote changed regions. In $\Omega_0$, SCD requires semantic consistency:
\begin{equation}
Y_{\mathrm{s1}}(p)=Y_{\mathrm{s2}}(p),
\quad p\in\Omega_0.
\end{equation}
SCP supports this property by activating $M_{\mathrm{nchg}}$. When $M_{\mathrm{nchg}}(p)$ is large, the two temporal features exchange information:
\begin{multline}
f'_{1,\mathrm{clean}}(p)
\approx
\phi_1
\left(
f'_{1,4}(p)
+
G_1(p)\odot f'_{2,4}(p)
\right),\\
p\in\Omega_0.
\end{multline}
This reduces temporal semantic discrepancy in unchanged regions. In $\Omega_1$, SCD requires transition discrimination:
\begin{equation}
Y_{\mathrm{scd}}(p)=
\Psi(Y_{\mathrm{s1}}(p),Y_{\mathrm{s2}}(p)),
\quad p\in\Omega_1.
\end{equation}
When $M_{\mathrm{nchg}}(p)$ is small, cross-temporal mixing is suppressed:
\begin{equation}
\begin{aligned}
f'_{1,\mathrm{clean}}(p)
&\approx
\phi_1(f'_{1,4}(p)),\\
f'_{2,\mathrm{clean}}(p)
&\approx
\phi_2(f'_{2,4}(p)),
\quad p\in\Omega_1.
\end{aligned}
\end{equation}
Thus, the discrepancy required for from-to semantic transition reasoning is preserved. \textbf{To intuitively capture this spatial alignment behavior and establish its mathematical foundations, a rigorous theoretical analysis of the SCP module---modeled as a region-conditioned discrepancy control framework governed by contractive and margin-preserving Lipschitz boundaries---is comprehensively documented in Section 1-1.5 of the Supplementary Material.}

MCE constructs multi-scale change evidence from the clean features:
\begin{equation}
\begin{aligned}
D_{\mathrm{clean}}
&=
|f'_{1,\mathrm{clean}}-f'_{2,\mathrm{clean}}|,\\
D_s
&=
\mathrm{Pool}_s(D_{\mathrm{clean}}),
\quad
B_s
=
\mathrm{Pool}_s(F_{\mathrm{BCE}}),
\quad s\in\{1,2,4\},\\
E_s
&=
B_s
\odot
\sigma(\phi_s(D_s)).
\end{aligned}
\end{equation}
The scale-wise evidence is fused and added to the initial BCE feature:
\begin{equation}
\begin{aligned}
F_{\mathrm{MCE}}
&=
\phi_{\mathrm{mce}}
\left(
E_1,\mathcal{U}(E_2),\mathcal{U}(E_4)
\right),\\
F_{\mathrm{FeaCE}}
&=
F_{\mathrm{MCE}}
+
F_{\mathrm{BCE}}.
\end{aligned}
\end{equation}

MCE provides scale-aware robustness. Let the clean difference contain a true semantic transition component $T$ and a nuisance component $\epsilon$:
\begin{equation}
D_{\mathrm{clean}}=T+\epsilon.
\end{equation}
For local independent nuisance, $s\times s$ average pooling reduces the variance approximately as
\begin{equation}
\mathrm{Var}(\mathrm{Pool}_{s}(\epsilon))
\approx
\frac{1}{s^2}
\mathrm{Var}(\epsilon).
\end{equation}
Therefore, the $2\times$ and $4\times$ branches attenuate local pseudo-change responses caused by illumination variation, seasonal noise, and slight registration error. The identity branch preserves fine boundaries and small changed objects. The residual connection with $F_{\mathrm{BCE}}$ prevents multi-scale fusion from removing detailed change cues.

Overall, FeaCE serves as a refinement stage after cross-temporal alignment. BCE extracts the initial temporal discrepancy, SCP uses the change and non-change masks to separate consistency refinement from transition preservation, and MCE aggregates change evidence across scales. Therefore, FeaCE does not define the temporal relation by itself. It strengthens the aligned representation produced by M-TBTT.

Consequently, the architectural workflow of SemDINO naturally revolves around M-TBTT. First, PyFu establishes the foundation by delivering semantic-prior-enhanced feature pyramids. Upon this input basis, M-TBTT executes bidirectional cross-temporal semantic alignment to yield mutually calibrated temporal states. The FeaCE pipeline then refines these aligned representations, successfully suppressing pseudo-changes while amplifying genuine semantic transitions. Finally, the ChangeFusion module and the decoupled SCD head translate the consolidated features into a CD map, S1/S2 semantic maps, and an edge map, which jointly compose the final SCD results. This symbiotic design establishes SemDINO as a fundamentally alignment-driven SCD framework, rather than a conventional change-mask refinement model with a superficially appended semantic branch.

\section{Experimental Results and Analysis}

\subsection{Experiments Settings}
\textbf{Datasets.}
The Landsat-SCD dataset \cite{Yuan_et_al._2022} is established based on Landsat-series remote sensing images acquired in Tumushuke, Xinjiang (39$^\circ$39$^\prime$–40$^\circ$04$^\prime$ N, 78$^\circ$53$^\prime$–79$^\circ$19$^\prime$ E) from 1990 to 2020. \textit{Geographically, the study area borders the Taklamakan Desert and features an extremely fragile ecological system, serving as a pivotal regional node of the Belt and Road economic corridor}. The dataset contains a total of 8,468 remote sensing image pairs with a consistent spatial resolution of 30 meters and a unified pixel dimension of 416 $\times$ 416 pixels. The land cover annotation system defines five major semantic categories, including cropland, desert, built-up areas, water bodies, and unchanged land areas. To guarantee the authenticity and reliability of experimental data, all samples produced by spatial interpolation augmentation were eliminated, yielding 2,425 original and unprocessed image pairs. These authentic samples were randomly partitioned into training, validation, and test subsets following a standard 6:2:2 division ratio \cite{Ding_et_al._2024}, \cite{Tang_et_al._2025}, corresponding to 1,455 training pairs, 485 validation pairs, and 485 test pairs respectively.

The SECOND dataset \cite{Yang_et_al._2021} is a large-scale publicly available benchmark for semantic change detection, which covers multiple urban areas across China with elaborate manual annotations. All images in this dataset are 512 $\times$ 512 pixels, with spatial resolutions varying from 0.5 m to 3 m. The dataset defines seven land-cover categories: unchanged areas, low vegetation, non-vegetated surface, trees, water bodies, buildings, and playgrounds, containing a total of 4662 image pairs. Following the official partitioning strategy in existing studies \cite{Yang_et_al._2021}, \cite{Tang_et_al._2025}, the original dataset assigns 2968 pairs for training and 1694 pairs for testing. In this work, we further subdivide the original test set to construct a validation subset, finally forming a new data division of 2968 training pairs, 847 validation pairs, and 847 test pairs.

The HRSCD dataset \cite{Daudt_et_al._2019} serves as a large-scale benchmark for semantic change detection. It consists of 291 pairs of 10,000 × 10,000 RGB aerial images captured from 2006 to 2012, covering a wide range of urban and rural landscapes. Five distinct land-cover categories are annotated within the dataset: artificial surfaces, agricultural areas, forests, wetlands, and water bodies. To reduce computational overhead, we split the original ultra-high-resolution imagery into non-overlapping 256 × 256 patches. Patches with a change ratio lower than 5\% are automatically filtered out, as they contain negligible change information. The retained image patches are randomly partitioned into training and test sets following an 8:2 split ratio.

WHU-CD \cite{Ji_et_al._2018}: This dataset is dedicated to building change detection, comprising aerial images of the same area captured before and after an earthquake. The original single image pair has a spatial resolution of 0.3 m, and we follow the standard experimental setup by cropping it into 256 × 256 patches. The widely adopted split results in 4,536 training pairs, 504 validation pairs, and 2,760 test pairs.

LEVIR-CD \cite{Chen_et_al._2020}: This dataset focuses on building change detection, containing Google Earth image pairs collected across 20 different regions with temporal intervals ranging from 5 to 14 years. The spatial resolution is 0.5 m, and the standard split yields 7,120 training pairs, 1,024 validation pairs, and 2,048 test pairs. Following common practice, we tile the images into non-overlapping 256 × 256 patches without additional curation.

\textbf{Implementation Details.}
All experiments are implemented based on PyTorch and conducted on NVIDIA RTX 3090 and Tesla A100 GPUs. For training optimization, distinct training configurations are adopted to accommodate the unique characteristics of different datasets. Specifically, on the Landsat-SCD dataset, we employ the stochastic gradient descent (SGD) optimizer with a momentum of 0.9 and a weight decay of $5\times10^{-4}$, where the initial learning rate is set to 0.01 and the batch size is set to 2. The learning rate is dynamically updated following the polynomial decay strategy with a power of 1.5 over 200 training epochs. Conversely, for the SECOND and HRSCD datasets, the Adam optimizer is utilized with a base learning rate of $1\times10^{-4}$ and a batch size of 8, also training for 200 epochs to ensure adequate convergence.

The entire framework is optimized by a composite loss function, which consists of cross-entropy loss for bi-temporal semantic segmentation, binary cross-entropy loss for change detection, change similarity loss for feature constraint, and auxiliary edge loss with a weight of 0.1 to strengthen boundary perception.

\textbf{Evaluation.} To quantitatively evaluate the performance of the proposed semantic change detection framework, we adopt five widely used evaluation metrics: binary change detection IoU (BCD-IoU), binary change detection F1-score (BCD-F1), separated Kappa (SeK), mean intersection over union (mIoU), and SCD F1-score ($F_{\text{scd}}$). The mIoU and SCD-related metrics ($F_{\text{scd}}$, SeK) focus on semantic-level change discrimination, while BCD-IoU and BCD-F1 are used to separately measure the binary change/unchange segmentation capability.

The detailed calculation formulas for SCD task's metrics are given as follows.

\begin{align}
\text{mIoU}           &= \frac{\text{IoU}_{\text{nc}} + \text{IoU}_{\text{c}}}{2}, \\
\text{IoU}_{\text{nc}} &= \frac{q_{0,0}}{\left(\sum_{i=0}^{C} q_{i,0} + \sum_{j=0}^{C} q_{0,j} - q_{0,0}\right)}, \notag \\
\text{IoU}_{\text{c}}  &= \frac{\sum_{i=1}^{C} \sum_{j=1}^{C} q_{i,j}}{\left(\sum_{i=0}^{C} \sum_{j=0}^{C} q_{i,j} - q_{0,0}\right)}. \notag \\
F_{\text{scd}}        &= \frac{2 \times P_{\text{scd}} \times R_{\text{scd}}}{P_{\text{scd}} + R_{\text{scd}}}, \\ 
P_{\text{scd}}        &= \frac{\sum_{i=1}^{C} q_{i,i}}{\sum_{i=1}^{C} \sum_{j=0}^{C} q_{i,j}}, \notag \\
R_{\text{scd}}        &= \frac{\sum_{i=1}^{C} q_{i,i}}{\sum_{i=0}^{C} \sum_{j=1}^{C} q_{i,j}}. \notag 
\end{align}

Different from standard Kappa coefficient, separated Kappa (SeK) focuses on the classification performance within changed regions. We construct a new confusion matrix $\widetilde{\mathbf{Q}} = \{\widetilde{q}_{i,j}\}$ by removing the category of unchanged pixels. Let $\rho$ be the overall accuracy on changed regions and $\eta$ be the expected accuracy under random prediction. Following the improved SeK calculation scheme in existing SCD researches, we further introduce an exponential weighting term associated with $\text{IoU}_\text{c}$ to enhance evaluation rationality. The corresponding formulas are defined as:

\begin{align}
\text{SeK} &= e^{\text{IoU}_{\text{c}} - 1} \cdot \frac{\rho - \eta}{1 - \eta}, \\
\rho &= \frac{\sum_{i=0}^{C-1} \widetilde{q}_{i,i}}{\sum_{i=0}^{C-1} \sum_{j=0}^{C-1} \widetilde{q}_{i,j}}, \nonumber \\
\eta &= \frac{\sum_{i=0}^{C-1} \left(\sum_{j=0}^{C-1} \widetilde{q}_{i,j} \times \sum_{j=0}^{C-1} \widetilde{q}_{j,i}\right)}{\left(\sum_{i=0}^{C-1} \sum_{j=0}^{C-1} \widetilde{q}_{i,j}\right)^2}. \nonumber 
\end{align}

\begin{table*}[htbp]
\centering
\setlength{\tabcolsep}{2.0pt}
\renewcommand{\arraystretch}{1.05} 
\captionsetup{aboveskip=5pt}
\caption{Comparison results for SCD task based on Landsat-SCD, SECOND and HRSCD datasets. Color brown denotes the best \textcolor{brown}{\textbf{values}} and the second-best \textbf{values} are in bold. }
\small
\label{tab:com1}
\begin{tabular}{l|lll|lll|lll}
\multirow{2}{*}{Methods} & \multicolumn{3}{l|}{\ \ \ \ \ \ \ \ \ \ \ \textbf{Landsat-SCD}} & \multicolumn{3}{l|}{\ \ \ \ \ \ \ \ \ \ \ \ \ \textbf{SECOND}} & \multicolumn{3}{l}{\ \ \ \ \ \ \ \ \ \ \ \ \ \ \ \textbf{HRSCD}}\\
 & \ \textit{mIoU (\%)} & \ \textit{Sek$_{17}$ (\%)} & \ \textit{$F_{scd}$ (\%)}        & \ \textit{mIoU (\%)}       & \ \textit{Sek$_{37}$ (\%)}    & \ \textit{$F_{scd}$ (\%)}        & \ \textit{mIoU (\%)}       & \ \textit{Sek$_{26}$ (\%)}    & \ \textit{$F_{scd}$ (\%)}    \\ 
\cmidrule[0.5pt]{1-10}
HRSCD4 \cite{Daudt_et_al._2019} & \ \ 79.19 & \ \ 32.27 & \ \ 73.21  & \ \ \ 71.08 & \ \ \ 17.71 & \ \ 59.01   & \ \ \ 63.37 & \ \ \ 4.94 & \ \ 49.64\\
SSCDl \cite{Ding_et_al._2022}  & \ \ 81.56 & \ \ 40.77 & \ \ 80.00  & \ \ \ 73.23 & \ \ \ 21.86 & \ \ 63.13   & \ \ \ 67.07 & \ \ \ 18.17 & \ \ 67.58\\
BiSRNet \cite{Ding_et_al._2022}  & \ \ 82.59 & \ \ 43.55 & \ \ 81.70  & \ \ \ 73.26 & \ \ \ 21.66 & \ \ 63.25   & \ \ \ 66.95 & \ \ \ 18.88 & \ \ 68.68\\
TED \cite{Ding_et_al._2024}  & \ \ 85.60 & \ \ 50.34 & \ \ 84.52  & \ \ \ 73.27 & \ \ \ 22.05 & \ \ 63.21   & \ \ \ 66.68 & \ \ \ 17.46 & \ \ 66.97\\
SCanNet \cite{Ding_et_al._2024}  & \ \ 85.50 & \ \ 50.77 & \ \ 85.16  & \ \ \ 73.43 & \ \ \ 22.27 & \ \ 63.82   & \ \ \ 68.28 & \ \ \ 20.86 & \ \ 70.53\\
DEFO-MLTSCD \cite{Li_et_al._2024}  & \ \ 87.79 & \ \ 56.38 & \ \ 87.21  & \ \ \ \textbf{73.71} & \ \ \ 22.71 & \ \ 63.65   & \ \ \ 66.58 & \ \ \ 17.54 & \ \ 67.16\\
MambaSCD \cite{Chen_et_al._2024}  & \ \ 85.57 & \ \ 51.50 & \ \ 85.30  & \ \ \ 73.47 & \ \ \ 22.86 & \ \ 63.97   & \ \ \ \textbf{68.82} & \ \ \textbf{\ 22.09} & \ \ \textbf{71.44}\\
LSAFNet \cite{Zhou_et_al._2024}  & \ \ 88.00 & \ \ 58.44 & \ \ 88.12  & \ \ \ 73.66 & \ \textbf{\ \ 23.13} & \ \ \textbf{64.52}   & \ \ \ 67.34 & \ \ \ 19.64 & \ \ 69.48\\
BT-SCD \cite{Tang_et_al._2025}  & \ \ \textbf{88.74} & \ \textbf{\ 59.32} & \ \ \textbf{88.56}  & \ \ \ 73.51 & \ \ \ 23.09 & \ \ 63.42   & \ \ \ 66.99 & \ \ \ 18.40 & \ \ 67.54\\
\cmidrule[0.5pt]{1-10}
\cellcolor{tong}\textbf{SemDINO (Ours)}  & \cellcolor{tong}\textcolor{brown}{\textbf{\ \ 89.51}} & \cellcolor{tong}\textcolor{brown}{\textbf{\ \ 62.24}} & \cellcolor{tong}\textcolor{brown}{\textbf{\ \ 89.92}}  & \cellcolor{tong}\textcolor{brown}{\textbf{\ \ \ 74.19}} & \cellcolor{tong}\textcolor{brown}{\textbf{\ \ \ 24.01}} & \cellcolor{tong}\textcolor{brown}{\textbf{\ \ 65.27}}   & \cellcolor{tong}\textcolor{brown}{\textbf{\ \ \ 69.23}} & \cellcolor{tong}\textcolor{brown}{\textbf{\ \ \ 23.39}} & \cellcolor{tong}\textcolor{brown}{\textbf{\ \ 71.84}}\\
\end{tabular}
\end{table*}

\subsection{Comparison Experiments and Analysis.}
\textbf{Results of Qualitative Analysis.} As reported in Table~\ref{tab:com1}, we evaluate SemDINO against nine SOTA methods across three benchmarks. In this section, we used the same data split and training parameters as in the method described in this paper to ensure a fair comparison. Remarkably, SemDINO achieves the best performance across all datasets and metrics, consistently outperforming CNN, Transformer, and Mamba-based baselines.

Specifically, on Landsat-SCD, SemDINO reaches an mIoU of $89.51\%$, a Sek of $62.24\%$, and an $F_{scd}$ of $89.92\%$. Compared to the strongest competitor BT-SCD \cite{Tang_et_al._2025}, our method achieves a significant performance improvement, with a $+0.92\%$ increase in the core semantic distinction metric, Sek. Meanwhile, early CNN frameworks like HRSCD4 \cite{Daudt_et_al._2019} lag behind SemDINO by nearly $30\%$ in Sek, exposing the inherent limitations of standard CNNs in capturing long-range cross-temporal dependencies.

Furthermore, on the high-resolution SECOND benchmark, SemDINO establishes a new SOTA baseline ($74.19\%$ mIoU, $24.01\%$ Sek, and $65.27\%$ $F_{scd}$), outperforming DEFO-MLTSCD \cite{Li_et_al._2024} in mIoU and LSAFNet \cite{Zhou_et_al._2024} in Sek and $F_{scd}$. Crucially, on the challenging, class-unbalanced HRSCD dataset, SemDINO consistently breaks through the performance ceiling, surpassing the advanced MambaSCD \cite{Chen_et_al._2024} across all metrics, achieving an mIoU of $69.23\%$, a Sek of $23.39\%$, and an $F_{scd}$ of $71.84\%$.

This consistent superiority demonstrates the architectural merits of our pipeline. While contemporary Transformer and Mamba models deliver competitive results, they often suffer from asymmetric unidirectional temporal alignment and neglect rich open-world semantic priors. SemDINO effectively overcomes these limitations by harmonizing DINO-driven dense semantic fusion, M-TBTT's symmetric bidirectional alignment, and a dual-consistency change enhancement mechanism, establishing a competitive baseline for high-precision semantic change detection.

\begin{figure*}[htbp]
    \centering
    \includegraphics[width=\textwidth]{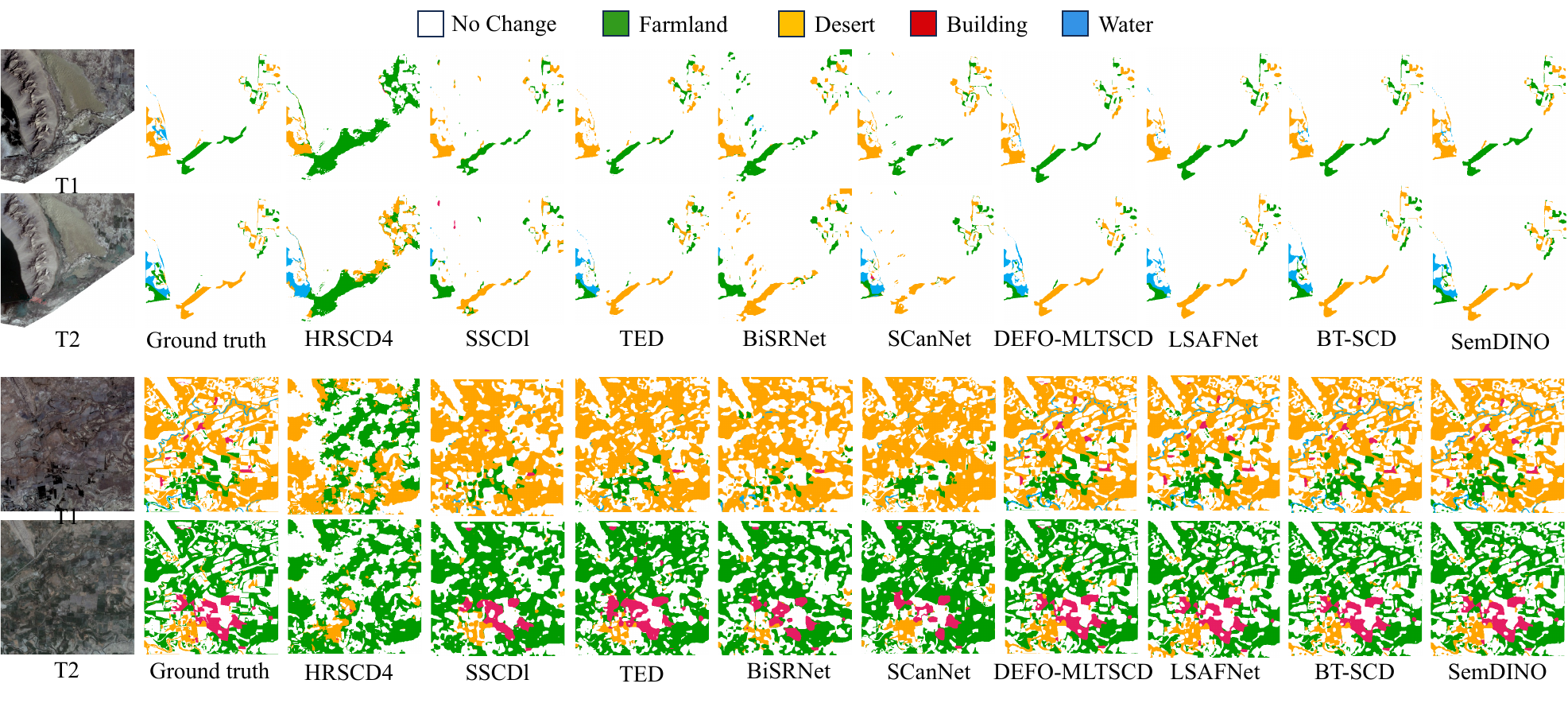}
    \caption{\small Visualization results on the Landsat-SCD datasets. (Zoom in to find more details.)}
\label{fig:vis-landsat}
\end{figure*}

\begin{figure*}[htbp]
    \centering
    \includegraphics[width=\textwidth]{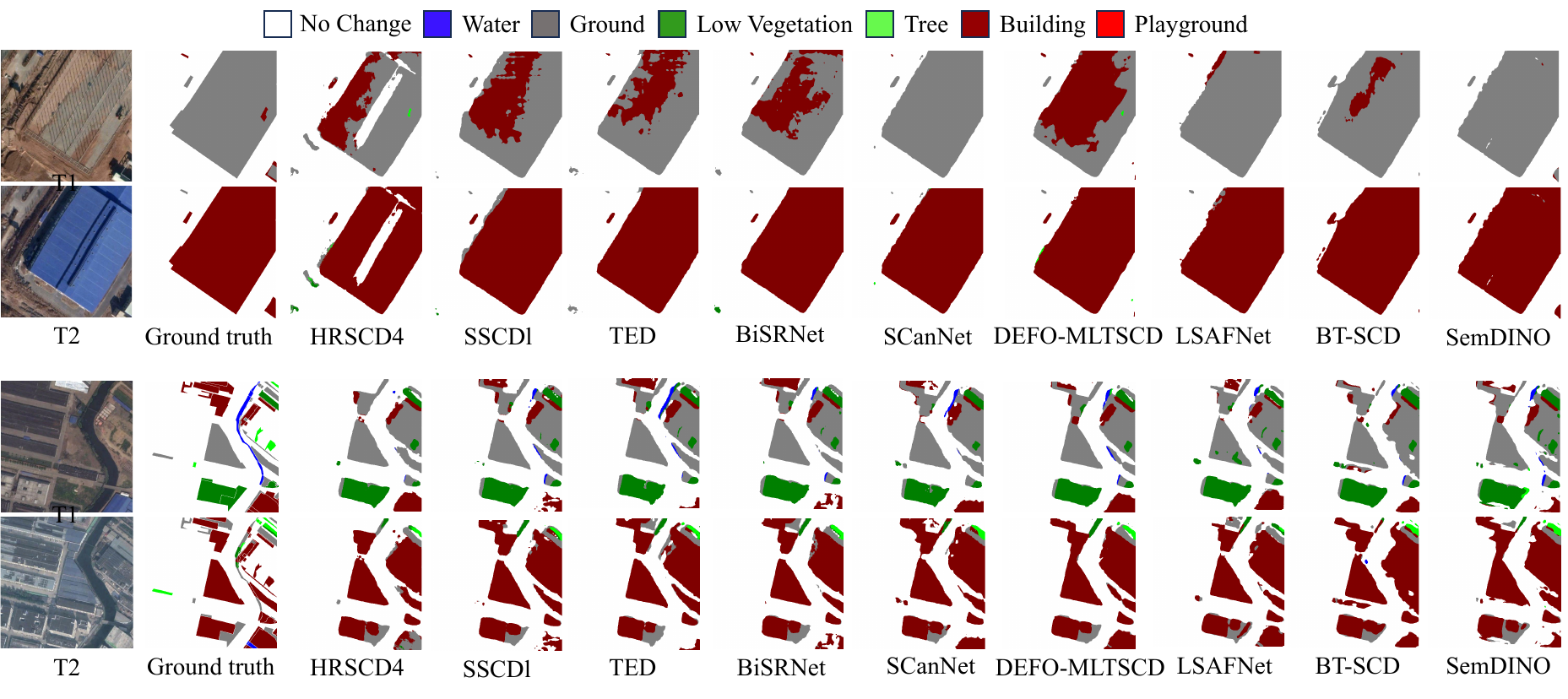}
    \caption{\small Visualization results on the SECOND datasets. (Zoom in to find more details.)}
\label{fig:vis-second}
\end{figure*}

\begin{figure*}[htbp]
    \centering
    \includegraphics[width=\textwidth]{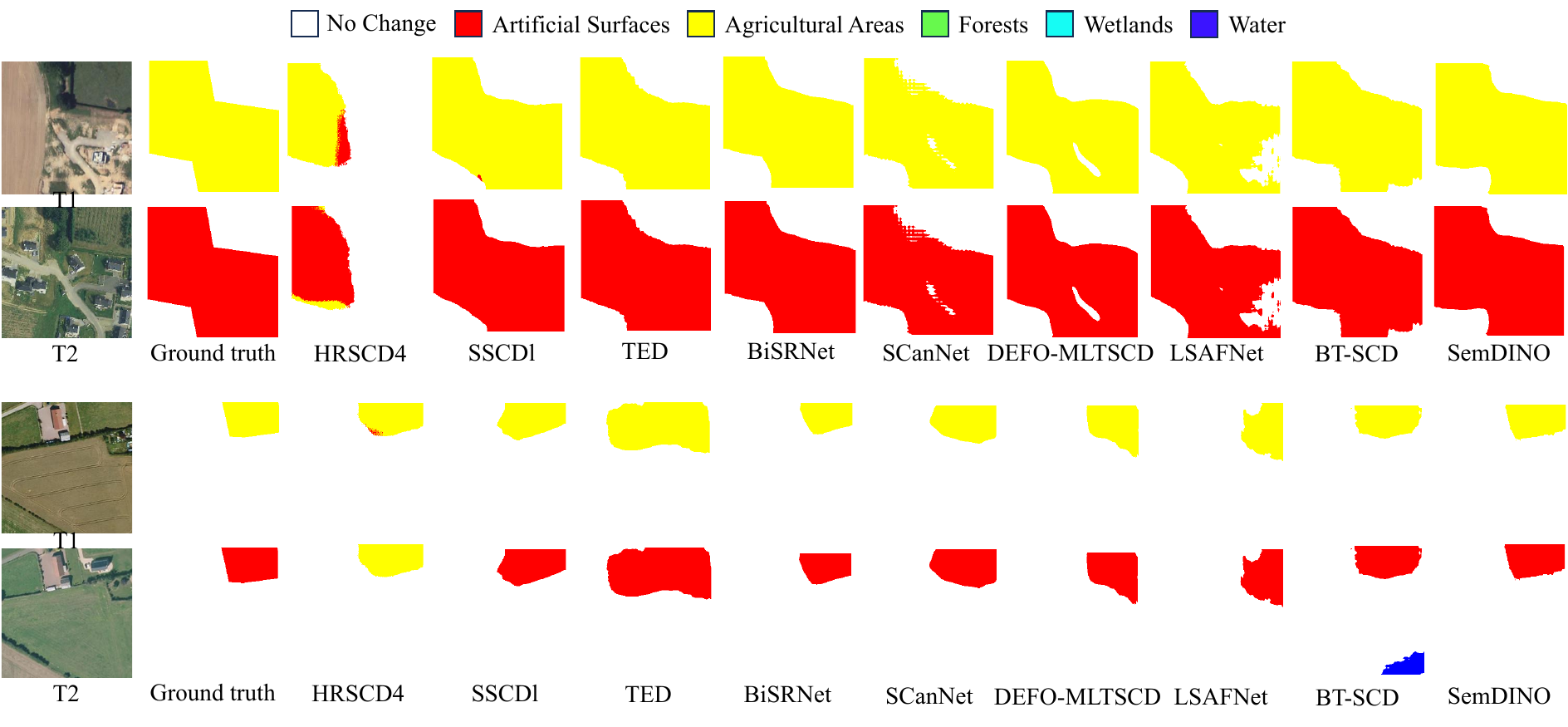}
    \caption{\small Visualization results on the HRSCD datasets. (Zoom in to find more details.)}
\label{fig:vis-hrscddataset}
\end{figure*}

\textbf{Results analysis of Visualization.} To qualitatively evaluate the multi-task land-cover transition parsing performance, we illustrate the comprehensive visual comparison maps across three benchmark datasets (Landsat-SCD, SECOND, and HRSCD) in Fig.~\ref{fig:vis-landsat} -~\ref{fig:vis-hrscddataset}, respectively. Globally, the comparative results exhibit a substantial alignment between the ground-truth annotations and the prediction maps generated by our SemDINO framework. Our model demonstrates remarkable superiorities in suppressing spatial clutter and producing razor-sharp boundary geometric delineations, consistently surpassing other state-of-the-art frameworks.

Specifically, as depicted in the Landsat-SCD visualization results (Fig.~\ref{fig:vis-landsat}), standard CNN-based models (such as HRSCD4 and BiSRNet) suffer severe land-cover category misclassifications and spatial fragmentation when handling challenging natural surface transitions, e.g., the complex temporal interactions among Farmland, Desert, and Water. For instance, in the second row of Fig.~\ref{fig:vis-landsat}, most baseline methods generate prominent false alarms (misinterpreting No Change backgrounds as changed regions) or fail to accurately track the fine-grained edge transitions of Buildings (red blocks). In sharp contrast, our SemDINO leverages the robust open-world dense semantic priors from the pre-trained foundation pipeline, producing highly integrated semantic change masks and achieving demonstrating enhanced pixel-level semantic consistency.

Furthermore, on the high-resolution SECOND dataset (Fig.~\ref{fig:vis-second}), tracking land-cover transition margins poses an immense challenge due to severe cross-temporal perspective distortions and shadow drift. In the complex urban block sample (the fourth row of Fig.~\ref{fig:vis-second}), previous methods like TED and SCanNet either output blurred boundary structures or generate chaotic semantic fragments among Ground, Low Vegetation, and Buildings. Benefiting from M-TBTT's bidirectional alignment paradigm, SemDINO establishes precise cross-temporal spatial correlations, yielding highly cohesive mask layouts that accurately restore complex geometric contours without being distracted by seasonal illumination drifts.

Lastly, the comparative visualizations on the HRSCD dataset (Fig.~\ref{fig:vis-hrscddataset}) further substantiate our framework's stability under highly unbalanced change distributions. In the dual-temporal agricultural plot scenario (the second row of Fig.~\ref{fig:vis-hrscddataset}), while contemporary approaches like LSAFNet and BT-SCD either introduce severe missing detections (false negatives) or misclassify Artificial Surfaces as Agricultural Areas, SemDINO cleanly decouples change boundaries and correctly maps out the semantic change trajectories. These visual patterns consistently reveal that by bridging deep foundation priors with fine-grained cross-temporal feature calibration, SemDINO demonstrates robust semantic generalization and boundary precision across diverse geographic contexts.

\subsection{Ablation Studies}
Ablation experiments were conducted on the proposed modules and strategies to validate their effectiveness. Table \ref{Table:Abla-encoder} to \ref{Table:Abla-feace} constitute the corresponding ablation experiments we conducted, with a detailed analysis as follows.

\begin{table*}[htbp]
\centering
\caption{Ablation studies of Encoder on Landsat-SCD. All metrics are expressed in percentage (\%) (a) Internal Ablation of the Pyramid Fusion (PyFu) Module. \textbf{Ours: Gated Fusion.} (b) Feature-scale Ablation. \textbf{Ours: Baseline (CNN) + DINOv3 + Adapter + PyFu.} }
\label{Table:Abla-encoder}
\setlength{\tabcolsep}{8.0pt}
\renewcommand{\arraystretch}{1.05}
\resizebox{\textwidth}{!}{
\begin{tabular}{@{}cc@{}}
\begin{minipage}[t]{0.39\textwidth}
\centering
\textbf{(a)} 
\begin{tabular}{l|lll}
\textbf{PyFu}        & \textit{mIoU}       & \textit{\ Sek}    & \textit{$F_{scd}$}  \\ 
\cmidrule[0.5pt]{1-4}
\cellcolor{tong}\textbf{Ours}  & \cellcolor{tong}\textcolor{brown}{\textbf{89.51}}   & \cellcolor{tong}\textcolor{brown}{\textbf{62.24}} & \cellcolor{tong}\textcolor{brown}{\textbf{89.92}} \\ 
Only CA   & 89.42  & 61.58 & 89.49 \\
Only SA   & 85.24  & 50.10 & 85.06 \\ 
Direct Concat  & 54.83  & 4.25 & 32.16 \\
\end{tabular}
\end{minipage}
&
\begin{minipage}[t]{0.45\textwidth}
\centering
\textbf{(b)} 
\begin{tabular}{l|lll}
\textbf{Feature Extraction}       & \textit{mIoU}       & \textit{\ Sek}    & \textit{$F_{scd}$}  \\ 
\cmidrule[0.5pt]{1-4}
Baseline (CNN)   & 88.82  & 61.12 & 89.05 \\ 
+DINOv3   & 89.18  & 61.72 & 89.50\\
+DINOv3+Adapter   & 89.38  & 61.98 & 89.70 \\
\cellcolor{tong}\textbf{Ours}  & \cellcolor{tong}\textcolor{brown}{\textbf{89.51}}  & \cellcolor{tong}\textcolor{brown}{\textbf{62.24}} & \cellcolor{tong}\textcolor{brown}{\textbf{89.92}} \\ 
\end{tabular}
\end{minipage}
\end{tabular}
}
\end{table*}

\begin{table*}[htbp]
\centering
\caption{Progressive ablation of M-TBTT on Landsat-SCD. All metrics are expressed in percentage (\%) (a) Internal structure ablation of the M-TBTT Module. \textbf{Ours: LG-g + Head = 4.} (b) Directional ablation of the M-TBTT module. \textbf{Ours: Multi-scale Bidirectional Temporal Transformer.}}
\label{Table:Abla-mtbtt}
\setlength{\tabcolsep}{3.0pt}
\renewcommand{\arraystretch}{1.05}
\resizebox{\textwidth}{!}{
\begin{tabular}{@{}cc@{}}

\begin{minipage}[t]{0.39\textwidth}
\centering
\textbf{(a)} 
\setlength{\tabcolsep}{7.0pt}
\renewcommand{\arraystretch}{1.15}
\begin{tabular}{l|lll}
\textbf{Structure}       & \textit{mIoU}       & \textit{\ Sek}    & \textit{$F_{scd}$}  \\ 
\cmidrule[0.5pt]{1-4}
\cellcolor{tong}\textbf{Ours}    & \cellcolor{tong}{\textcolor{brown}{\textbf{89.51}}}  & \cellcolor{tong}\textcolor{brown}{\textbf{\textcolor{brown}{62.24}}} & \cellcolor{tong}\textcolor{brown}{\textbf{\textcolor{brown}{89.92}}} \\ 
w/o LG-g   & 89.44  & 61.69 & 89.54 \\
Head = 1   & 89.37  & 61.45 & 89.45 \\
Head = 2   & 89.42  & 61.59 & 89.51 \\ 
\end{tabular}
\end{minipage}
&
\begin{minipage}[t]{0.48\textwidth}
\centering
\textbf{(b)} 
\begin{tabular}{l|lll|lll}
\ \ \textbf{Direction}  &  T1$\rightarrow$T2  & T1$\leftarrow$T2 & T1$\leftrightarrow$ T2        & \textit{mIoU}       & \textit{\ Sek}    & \textit{$F_{scd}$}  \\ 
\cmidrule[0.5pt]{1-7}
w/o Forward  & $\times$  & $\checkmark$ & $\times$  & 89.39 & 61.47 & 89.46\\ 
w/o Reverse  & $\checkmark$  & $\times$ & $\times$  & 89.08 & 60.53 & 89.08\\
\cmidrule[0.5pt]{1-7}
\cellcolor{tong}\textbf{Ours}  & \cellcolor{tong}$\checkmark$  & \cellcolor{tong}$\checkmark$ & \cellcolor{tong}$\checkmark$  & \cellcolor{tong}{\textbf{\textcolor{brown}{89.51}}} & \cellcolor{tong}\textcolor{brown}{\textbf{62.24}} & \cellcolor{tong}\textcolor{brown}{\textbf{89.92}}\\
w/o M-TBTT & $\times$  & $\times$ & $\times$  & 89.41 & 61.55 & 89.49\\
\end{tabular}
\end{minipage}
\end{tabular}
}
\end{table*}

\textbf{Effectiveness of SemDINO's Encoder.} We multi-dimensionally investigate our encoder components on the Landsat-SCD dataset. As shown in Table~\ref{Table:Abla-encoder}(a), the necessity of our gated fusion in the PyFu module is intensely justified. Under the ``Direct Concat'' setup where the frozen DINOv3 priors and CNN features are naively concatenated without attention-guided filtering, a notable accuracy drop occurs, with $mIoU$ and $Sek$ sharply dropping to $54.83\%$ and $4.25\%$. This failure empirically confirms that the significant cross-modal domain gap can heavily corrupt change representations by unconditionally injecting raw features. Conversely, while independent attention pathways (``Only CA'' or ``Only SA'') provide progressive metric increments, our full dual-gating configuration yields the optimal performance. This validates our theoretical insights in Section I-D of the Supplementary Material: the joint spatial-channel gating wrapper effectively disciplines prior injection, facilitating robust adaptive semantic compensation while filtering out noise. Concurrently, Table~\ref{Table:Abla-encoder}(b) demonstrates the stepwise performance elevation across feature pipelines. The pure CNN baseline establishes an initial bottleneck ($mIoU=88.82\%$). Sequentially incorporating the frozen DINOv3 prior (``+DINOv3'') and its dedicated Separate Adaptation Blocks (``+DINOv3+Adapter'') systematically boosts $mIoU$ to $89.18\%$ and $89.38\%$, respectively, proving the inherent richness of the foundation model's dense semantics. Finally, complete coordination with the PyFu module pushes metrics to the maximum ($mIoU=89.51\%$, $Sek=62.24\%$). This progressive optimization underscores that SemDINO successfully balances local structure and global semantics, resolving the critical trade-off between semantic insufficiency and domain discrepancy.

\textbf{Effectiveness of M-TBTT.} We validate the multi-scale bidirectional temporal transformer (M-TBTT) through structural and directional ablation groups (Table~\ref{Table:Abla-mtbtt}). As captured in Table~\ref{Table:Abla-mtbtt}(a), both the multi-head mechanism and the learnable gate ($LG\text{-}g$) yield consistent gains. Metrics monotonically degrade as the attention head count decreases from 4 to 1, confirming that a multi-head design enhances multi-scale temporal modeling. Eliminating $LG\text{-}g$ also compromises performance, verifying its capability to adaptively weight bidirectional temporal residuals across change and non-change regions. Crucially, Table~\ref{Table:Abla-mtbtt}(b) reveals that disabling either single-directional attention performs worse than removing the entire M-TBTT module. This anomaly arises because an incomplete branch forces zero residual maps, causing the learnable gate to inject noisy gradients when aggregating zero tensors with original features. In contrast, the ``w/o M-TBTT'' variant preserves clean encoder features without temporal interference. 

Our full bidirectional design simultaneously captures expanding and vanishing land-cover alterations, yielding optimal performance. More importantly, by maintaining both interaction flows ($T_1\leftrightarrow T_2$) under strict parameter sharing, M-TBTT not only captures mutual cross-temporal dependencies but also eliminates temporal-order sensitivity during inference, ensuring mathematically symmetric and stable predictions regardless of temporal input ordering.

\begin{table}[htbp]
\centering
\small
\caption{Progressive ablation of \textbf{FeaCE} (Change Enhancement) on Landsat-SCD. All metrics are expressed in percentage (\%). \textbf{Ours: BCE + SCP + MCE.} }
\label{Table:Abla-feace}
\setlength{\tabcolsep}{4.5pt}  
\renewcommand{\arraystretch}{1.05} 
\begin{tabular}{l|lll|llll}
\ \ \ Setting  &  BCE  & SCP & MCE        & \textit{mIoU}       & \textit{\ Sek}    & \textit{$F_{scd}$}  \\ 
\cmidrule[0.5pt]{1-7}

w/. BCE   & $\checkmark$  & $\times$ & $\times$  & \ 89.36 & 61.42 & 89.45\\ 
w/. SCP  & $\times$  & $\checkmark$ & $\times$  & \ 89.10 & 60.68 & 89.18\\
w/. MCE  & $\times$  & $\times$ & $\checkmark$  & \ 89.38 & 61.49 & 89.47\\
w/o BCE  & $\times$  & $\checkmark$ & $\checkmark$  & \ 89.16 & 60.81 & 89.23\\
w/o SCP  & $\checkmark$  & $\times$ & $\checkmark$  & \ \textbf{89.45} & \textbf{62.01} & \textbf{89.60}\\
w/o MCE  & $\checkmark$  & $\checkmark$ & $\times$  & \ 89.36 & 61.36 & 89.42\\
\cmidrule[0.5pt]{1-7}
\cellcolor{tong}\textbf{Ours}  & \cellcolor{tong}$\checkmark$  & \cellcolor{tong}$\checkmark$ & \cellcolor{tong}$\checkmark$  & \cellcolor{tong}{\textbf{\textcolor{brown}{\ 89.51}}} & \cellcolor{tong}\textcolor{brown}{\textbf{62.24}} & \cellcolor{tong}\textbf{\textcolor{brown}{89.92}}\\
w/o FeaCE & $\times$  & $\times$ & $\times$  & \ 89.24 & 61.05 & 89.29\\
\end{tabular}
\end{table}

\textbf{Effectiveness of \#FeaCE.} We deploy a multi-stage Feature Change Enhancement (FeaCE) workflow—comprising BCE, SCP, and MCE—to suppress pseudo-variations induced by illumination, seasonality, and registration errors. Table~\ref{Table:Abla-feace} demonstrates the progressive contribution of each submodule. Specifically, removing BCE entirely severely degrades representation learning: without baseline difference features, SCP cannot extract valid change masks for semantic purification, and MCE lacks the necessary inputs for temporal enhancement. This full deprivation forces the remaining decoupled modules to introduce detrimental noise, resulting in performance inferior to the ``w/o FeaCE'' baseline. Similarly, retaining only SCP yields the worst performance due to the absence of temporal difference guidance for its gating operations. Conversely, when all three submodules corporate seamlessly, BCE establishes fundamental difference representation, SCP effectively eliminates pseudo-change noise, and MCE aggregates multi-scale enhanced features, achieving the optimal quantitative performance.

In summary, the ablation experiments systematically establish the structural necessity and functional synergy of SemDINO's architecture. Rather than a naive assembly of isolated components, the results demonstrate that robust spatial semantic priors (PyFu) and bidirectional temporal dynamics (M-TBTT) act as mutually reinforcing pillars: DINOv3's broad open-world knowledge drives noise-resistant cross-temporal alignment, while M-TBTT's symmetric calibration guides static spatial features into time-aware representations. Downstream components (FeaCE) seamlessly capitalize on these mutually calibrated representations to eliminate pseudo-changes, altogether proving that SemDINO's performance leap stems from an intrinsically coordinated, synergy-driven design.

\begin{table*}[htb]
\centering
\setlength{\tabcolsep}{10.0pt}
\renewcommand{\arraystretch}{1.05} 
\small
\caption{Comprehensive evaluation of SemDINO's generalization capabilities across various companion CNN module variants and alternative pre-trained Vision Foundation Model (VFM) backbones. $^\dagger$ denotes the frozen DINOv3 backbone. $^\ddagger$ denotes the trainable CNN module with ImageNet pre-training. Our SemDINO (DINOv3+ResNet50)}
\label{Table:Dis-pretrain}
\begin{tabular}{l|ccc}
\multirow{2}{*}{Methods} & mIoU & Sek  & \#Params \\
& (\%) & (\%) & \textit{(Pre-train-Enc.)+\textbf{w/o Enc.}} \\
\cmidrule[0.5pt]{1-4}
\cellcolor{tong}\textbf{DINOv3$^\dagger$ + ResNet50$^\ddagger$ (Ours)}  &\cellcolor{tong}\textcolor{brown}{\textbf{89.51}}&\cellcolor{tong}\textcolor{brown}{\textbf{62.24}}  & \cellcolor{tong}(230+25)M + 2.09M \\
DINOv3$^\dagger$ + ResNet34$^\ddagger$  &\textbf{89.11} &\textbf{61.75} &(230+21)M + 2.09M \\
DINOv3$^\dagger$ + MobileNetV2$^\ddagger$  &88.32 &60.92 & (230+3)M + 2.09M \\
DINOv3$^\dagger$ (w/o CNN)  &72.14 &21.89 & 230*M + 2.09M \\
\cmidrule[1.5pt]{1-4}
VFM Backbone Variant & mIoU & Sek  & Pre-training Paradigm \\
\cmidrule[0.5pt]{1-4}
w/o VFM (CNN)  &88.82 &61.12 & Supervised (ImageNet-1K) \\
w/ CLIP-Vision-L/14  &88.45 &60.54 & Contrastive (Text-Image) \\
w/ SAM (ViT-L/16)  &88.91 &58.76 & Promptable Segmentation \\
w/ DINOv2-L/16  &\textbf{89.24} &\textbf{61.85} & Self-supervised (Fea. Dis.) \\
\cellcolor{tong}\textbf{w/} \textbf{DINOv3-L/16 (Geo-adapted)}  &\cellcolor{tong}\textcolor{brown}{\textbf{89.51}} &\cellcolor{tong}\textcolor{brown}{\textbf{62.24}} & \cellcolor{tong} Remote Sensing Domain-adapted \\
w/ DINOv3-L/16 (Natural)  &\textbf{89.39} &\textbf{62.01} & Self-supervised (Open-world) \\
\end{tabular}
\end{table*}

\begin{table}[htbp]
\centering
\small
\caption{Generalization performance on BCD task. The DINOv3 backbone and intermediate MoFPN are fixed. only the prediction head is replaced for binary change detection.}
\label{Table:Dis_BCD}
\setlength{\tabcolsep}{6.0pt}
\renewcommand{\arraystretch}{1.05}
\begin{tabular}{l|cc|cc}
\multirow{2}{*}{Methods} & \multicolumn{2}{c|}{WHU-CD} & \multicolumn{2}{c}{LEVIR-CD}\\
& F1(\%) & IoU(\%) & F1(\%) & IoU(\%) \\
\cmidrule[0.5pt]{1-5}
IFNet \cite{Zhang_et_al._2020} & 89.82 & 81.52 & 91.60 & 84.51 \\
BIT \cite{Chen_et_al._2021} & 80.97 & 68.02 & 89.94 & 81.72\\
ChangeFormer \cite{Bandara_et_al._2022} & 87.96 & 78.51 & 89.92 & 81.69 \\
ChangeCLIP \cite{Dong_et_al._2024} & 90.05 & 81.91 & 92.04 & 85.26\\
CDMamba \cite{Zhang_et_al._2025b} & 91.13 & 83.71 & 90.34 & 82.38\\
ChangeDINO \cite{Cheng_et_al._2026} & \textcolor{brown}{\textbf{94.18}} & \textcolor{brown}{\textbf{89.00}} & \textbf{92.31} & \textbf{85.72}\\
\cmidrule[0.5pt]{1-5}
\cellcolor{tong}\textbf{SemDINO (Ours)} &\cellcolor{tong}\textbf{93.72} &\cellcolor{tong}\textbf{88.89} &\cellcolor{tong}\textcolor{brown}{\textbf{92.97}} &\cellcolor{tong}\textcolor{brown}{\textbf{86.51}} \\
\end{tabular}
\end{table}

\section{Discussion}
\textbf{Generalization and Robustness Testing.} In this section, we provide a systemic evaluation of SemDINO regarding its structural generalization capabilities across pre-trained backbones and downstream task variations, followed by a quantitative robustness stress-testing against real-world environmental pseudo-changes.

\subsection{Analysis of Generalization in Pre-trained Backbones}
To comprehensively evaluate the generalization scalability of our dual-stream encoder, we systematically conduct decoupling investigations from two orthogonal perspectives: the structural configuration of the companion trainable CNN stream and the algorithmic selection of the frozen Vision Foundation Model (VFM) dense semantic prior. All empirical results are horizontally aggregated in Table~\ref{Table:Dis-pretrain}.

Firstly, as quantified in the upper part of Table~\ref{Table:Dis-pretrain}, we freeze the DINOv3 backbone and sequentially evaluate four companion CNN branches: ResNet50 (our standard SemDINO), ResNet34, MobileNetV2, and a variant completely stripped of the CNN pathway. Experimental profiles indicate that the DINOv3+ResNet50 setup delivers the optimal metrics ($89.51\%$ mIoU and $62.24\%$ SeK). Downscaling to a shallower ResNet34 or a lightweight MobileNetV2 restricts representation learning due to shrunk multi-scale geometric capacity, while discarding the CNN branch entirely results in the lowest performance ($72.14\%$ mIoU). Notably, the standalone DINOv3 variant still maintains an acceptable prediction baseline, confirming its dominant role as a semantic anchor. This confirms that while frozen VFM features anchor general land-cover semantics, the trainable CNN stream supplements task-specific local structural details to resolve representation insufficiency.

Secondly, the lower part of Table~\ref{Table:Dis-pretrain} cross-compares alternative pre-trained VFM priors integrated into our network while keeping the downstream pipelines unaltered. Relying on CLIP-Vision-L/14 yields suboptimal gains ($88.45\%$ mIoU), given that image-text contrastive learning favors global cross-modal pairing rather than granular, dense pixel-level representations. The SAM backbone enhances change localization via structural edge prompts but severely lacks category-level multi-class discrimination, creating a bottleneck in semantic metrics ($58.76\%$ SeK). Conversely, DINOv3-L/16 consistently outperforms its predecessor DINOv2-L/16 ($89.24\%$ mIoU). Crucially, initializing with remote sensing domain-adapted weights (DINOv3-L/16 Geo-adapted) achieves superior performance over its open-world natural-image counterpart (DINOv3-L/16 Natural), improving mIoU and SeK by $+0.12\%$ and $+0.23\%$, respectively. This edge highlights that DINOv3's advanced multi-task self-supervised distillation paradigm, combined with large-scale satellite pre-training, endows the model with an inherent inductive bias for overhead spatial distributions and complex land-cover geometries, validating its employment as our core prior generator.

\subsection{Analysis of \textbf{SemDINO/CD-Head Module} Generalization}
To verify the generalization and flexibility of our SemDINO framework and prediction head, we validate its performance on the binary change detection (BCD) task. 
Specifically, the entire proposed SemDINO architecture remains unchanged, and only the CD-Head is flexibly switched to either an SCD-Head or a BCD-Head for different task requirements.

\begin{table*}[htbp]
\centering
\setlength{\tabcolsep}{5.0pt}
\renewcommand{\arraystretch}{1.20}
\captionsetup{aboveskip=5pt}
\caption{Quantitative robustness stress-testing under registration noise and illumination drift on Landsat-SCD dataset. \textcolor{brown}{\textbf{Brown}} text denotes the best values, while \textbf{bold} text highlights the second-best values. Values in parentheses represent performance degradation relative to the noise-free baseline. * indicates the highest mIoU under noise conditions.}
\small
\label{tab:robustness_stress}
\begin{tabular}{l|ccc|cc}
\multirow{2}{*}{Method} & \multicolumn{3}{c|}{Registration Displacement Noise ($\Delta p$ Shifting)} & \multicolumn{2}{c}{Illumination Drift ($\alpha$ Scaling)} \\
& $\Delta p = 1$ px & $\Delta p = 2$ px & $\Delta p = 3$ px & $\alpha = 0.8$ (Dim) & $\alpha = 1.2$ (Bright) \\
\midrule
HRSCD4 & 69.49\% (-9.7\%) & 66.86\% (-12.3\%) & 64.31\% (-14.9\%) & 70.73\% (-8.46\%)& 70.91\% (-8.28\%)\\
BiSRNet & 75.27\% (-7.3\%) & \textbf{73.06\% (-9.5\%)} & \textcolor{brown}{\textbf{70.46\% (-12.1\%)}} & 76.40\% (-6.19\%)& 76.77\% (-5.82\%)\\
TED & 79.28\% (-6.3\%) & 75.84\% (-9.8\%) & 71.66\% (-13.9\%) & 80.67\% (-4.93\%) & 80.96\% (-4.64\%)\\
DEFO-MLTSCD & 77.86\% (-9.9\%) & 74.24\% (-13.6\%) & 71.00\% (-16.8\%) & 80.39\% (-7.40\%) & 80.40\% (-7.39\%)\\
BTSCD & \textbf{84.31\% (-4.4\%)} & 78.82\% (-9.9\%) & \textbf{75.44\% (-13.3\%)} & \textbf{88.66\% (-0.08\%)} & \textbf{88.60\% (-0.14\%)} \\
\midrule
\cellcolor{tong}\textbf{SemDINO (Ours)} & \cellcolor{tong}\textcolor{brown}{\textbf{\ *85.43\% (-4.1\%)}} & \cellcolor{tong}\textcolor{brown}{\textbf{*80.13\% (-9.4\%)}} & \cellcolor{tong}\ *75.57\% (-13.9\%) & \cellcolor{tong}\textcolor{brown}{\textbf{\ *89.45\% (-0.06\%)}} & \cellcolor{tong}\textcolor{brown}{\textbf{*89.44\% (-0.07\%)}} \\
\end{tabular}
\end{table*}

Table \ref{Table:Dis_BCD} presents the experimental results on the WHU-CD and LEVIR-CD datasets for the BCD task. These results verify the plug-and-play property of our prediction head, demonstrating that the framework can be flexibly adapted to different change detection tasks by simply replacing the head, without modifying the core feature extraction and fusion pipeline.
This strong generalization confirms the high universality of the proposed SemDINO, enabling its broad applicability across various remote sensing change detection scenarios.

\subsection{Robustness Stress-Testing Methodology and Results}
To empirically verify robustness against registration misalignment and illumination drift, we introduce synthetic perturbations into the Landsat-SCD testing dataset:
\begin{enumerate}
    \item \textbf{Registration Displacement Noise:} Spatial pixel-shifting $\Delta p \in \{1, 2, 3\}$ pixels horizontally and vertically is applied to $T_2$ frames to simulate satellite tracking misalignment. 
    \item \textbf{Illumination and Contrast Drift:} Brightness and contrast scaling factors $\alpha \in \{0.8, 1.2\}$ are injected on $T_2$ images to mimic solar zenith and seasonal phenological variations.
\end{enumerate}

As documented in Table~\ref{tab:robustness_stress}, conventional semantic change detection frameworks exhibit acute vulnerability to spatial perturbations. When subjected to increasing spatial misalignments ($\Delta p = 1, 2, 3$ px), baseline models undergo severe accuracy degradation, with DEFO-MLTSCD dropping by up to $-16.8\%$ and HRSCD4 falling to $64.31\%$ at $\Delta p = 3$ px. This confirms that traditional architectures fail to disentangle synthetic pixel displacements from genuine geophysical transitions.

Conversely, our proposed SemDINO achieves the highest absolute mIoU across all noise intensity levels, demonstrating superior spatial resilience. Specifically, under minor to moderate misalignments ($\Delta p = 1$ px and $2$ px), SemDINO preserves remarkable performance ($85.43\%$ and $80.13\%$) while exhibiting the lowest degradation margins ($-4.1\%$ and $-9.4\%$, respectively). Even under the most extreme distortion ($\Delta p = 3$ px), SemDINO firmly retains the highest change mIoU of $75.57\%$, outperforming the strongest baseline BTSCD ($75.44\%$) and TED ($71.66\%$). This spatial stability is primarily credited to M-TBTT's bidirectional cross-temporal alignment paradigm and MCE's multi-scale pooling buffers, which cooperatively rectify feature mismatches caused by spatial shifting.

Furthermore, the right-hand columns of Table~\ref{tab:robustness_stress} validate the models' adaptability toward illumination drift. While contemporary models suffer apparent fluctuations under illumination scaling, SemDINO demonstrates near-perfect immunity to drastic radiometric shifts. Specifically, it achieves a stellar mIoU of $89.45\%$ under light attenuation ($\alpha = 0.8$) and $89.44\%$ under excessive brightness ($\alpha = 1.2$), suffering a negligible decay of less than $0.07\%$ relative to its clean benchmark ($89.51\%$). This remarkable stability substantiates that by harnessing extensive open-world dense semantic priors from the pre-trained foundation encoder, SemDINO captures high-level structural invariants rather than fragile low-level pixel radiances, confirming its profound generalizability against complex real-world environmental variations.

\section{Conclusion}

This work revisits semantic change detection from the perspective of cross-temporal semantic alignment. The proposed framework demonstrates that reliable semantic transitions depend on whether the representations of two temporal observations can be calibrated into a consistent semantic space before change reasoning. By integrating transferable visual priors with task-specific spatial representations and enforcing symmetric temporal interaction, the proposed approach provides a principled solution to the instability caused by temporal ordering, domain discrepancy, and environmental variations.

Beyond improving detection accuracy, this study highlights several implications for future remote sensing representation learning. First, foundation models should not be directly transferred as universal feature extractors, but should be selectively adapted according to the semantic and geometric requirements of downstream tasks. Second, temporal modeling in change analysis should consider the intrinsic symmetry between observations, since neither temporal state should be treated as a fixed reference when identifying semantic evolution. Third, change detection can benefit from separating semantic consistency preservation from transition discrimination, enabling models to distinguish meaningful land-cover evolution from incidental observation differences.

The proposed framework provides a general design perspective for future large-scale Earth observation systems, where robust change understanding requires the joint consideration of prior knowledge, temporal relationships, and semantic structures. Future studies may further explore adaptive alignment strategies under larger temporal intervals, heterogeneous sensing modalities, and open-world geographic scenarios to develop more generalizable models for dynamic environmental monitoring.

\end{document}